\documentclass[journal]{vgtc}                     % final (journal style)

\usepackage{amssymb}
\usepackage{newtxmath}
\usepackage{overpic}

\onlineid{0}

\vgtccategory{Research}

\vgtcpapertype{algorithm/technique}

\title{Measuring Distortion in the Empty Regions of Dimensionality Reduction Scatterplots with the Gap Index}
\author{%
  \authororcid{Jaume Ros}{0009-0003-9288-843X},
  \authororcid{Alessio Arleo}{0000-0003-2008-3651}, and
  \authororcid{Fernando Paulovich}{0000-0002-2316-760X}
}

\authorfooter{
  \item
  Jaume Ros, Alessio Arleo, and Fernando Paulovich are with Eindhoven University of Technology.\\E-mail: \{j.ros.alonso\,$|$\,a.arleo\,$|$\,f.paulovich\}@tue.nl\,.
}

\abstract{%
Quality metrics play a crucial role in the proper use of dimensionality reduction projections for visual analysis of high-dimensional data. They quantify the degree of distortion of a projection compared to the high-dimensional data and provide a reliable indication of how confident users can be in the structures they see in the resulting layouts. However, most popular metrics focus on capturing direct relationships between points (e.g., distances or neighborhoods) while neglecting distortions in empty areas of the layout, even though these often compose visually relevant features of a 2D layout. In this paper, we introduce the Gap Index (GI), a quality metric for 2D projections that captures visual distortion by measuring spatial distortion in empty areas of a projection. It does so by decomposing the space into empty triangles, which are then compared to their high-dimensional counterparts to compute the deformation. This per-triangle deformation can be aggregated into a single scalar value or overlaid on a projection to visualize regional distortion patterns. Results show that, contrary to popular quality metrics, the GI is sensitive to small structural deformations that have high visual impact. It is also fast to compute and interpretable.
}


\teaser{
  \centering
  \vspace{2em}
  \includegraphics[width=\linewidth, alt={Overview of the computation of the GI.}]{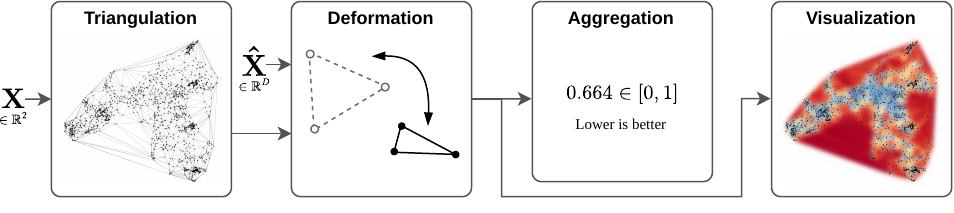}
  \caption{Overview of the computation of the Gap Index. Firstly, the projection is decomposed into triangles, which are then matched to their corresponding counterparts in high-dimensional space, and a deformation value is computed for each. The result can be aggregated into a single scalar quantifying the overall distortion, or overlaid on the scatterplot in order to see the distortion; the empty areas are colored with a divergent scale from blue (compression) to red (stretch), with yellow indicating no distortion.}
  \label{fig:teaser}
}

\begin{document}

\maketitle

%%%%%%%%%%%%%%%%%%%%%%%%%%%%%%%%%%%%%%%%%%%%%
%%%%%%%%%%%%%%%%%%%%%%%%%%%%%%%%%%%%%%%%%%%%%
%%%%%%%%%%%%%%%%%%%%%%%%%%%%%%%%%%%%%%%%%%%%%
\section{Introduction} \label{sec:introduction}

%
% general motivation and problem
%
Dimensionality Reduction (DR) plays a crucial role in the analysis of high-dimensional data. By projecting the data points into a two-dimensional space, the data can be visualized as a scatterplot, enabling a variety of visual analytics tasks~\cite{Nonato_Aupetit_2019}.
However, unless the original data lies on a two-dimensional manifold, any DR projection will necessarily entail some distortion. This can take the form of a loss of information about the high-dimensional structure that cannot be displayed in 2D, and the appearance of spurious patterns in the plot that are not present in the original data. Although this effect is known to DR experts, measuring the reliability of the projection is crucial for improving confidence in the visual analysis insights~\cite{mokbel2013visualizing}.

%
% How this is handled in the literature
%
Quality metrics serve to quantify the distortion introduced by the DR process. There are numerous examples that specialize in capturing specific properties or distortions in the data~\cite{Lee_Verleysen_2009, Jeon_Cho_2023}.
They are an essential part of any analysis that leverages DR. Without measuring the quality of the projection, it is more difficult to ``trust'' the patterns displayed by (and absent from) the projection.
Additionally, quality metrics can support visual analysis by providing feedback on which regions of the projection have experienced higher degrees of distortion~\cite{Lespinats_Aupetit_2011}.

\begin{figure}[t]
    \centering
    \begin{minipage}{0.49\columnwidth}
        \centering
        \includegraphics[height=3.8cm,keepaspectratio,alt={PCA projection of the plane dataset}]{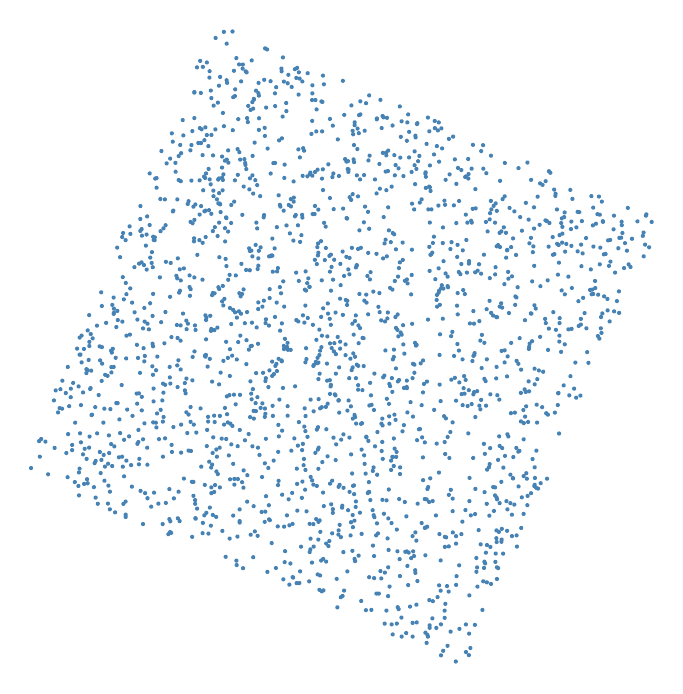}\\
        % \begin{overpic}[height=3.8cm,keepaspectratio]{figs/example_plane_PCA.pdf}
        %     \put(0,90){PCA}
        % \end{overpic}
        \small
        \textbf{(a) PCA}\\
        Stress ($\downarrow$): $0$\\
        Trustworthiness ($\uparrow$): $1$
    \end{minipage}%
    \hfill
    \begin{minipage}{0.49\columnwidth}
        \centering
        \includegraphics[height=3.8cm,keepaspectratio,alt={t-SNE projection of the plane dataset}]{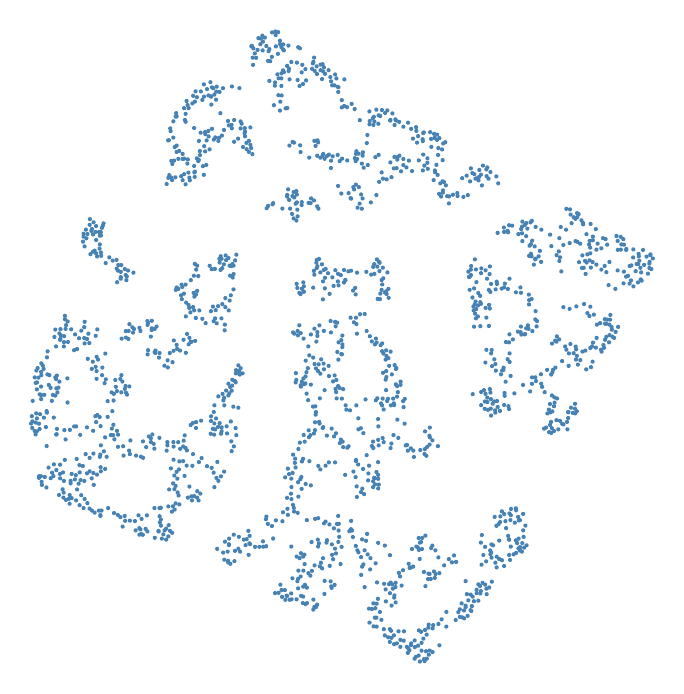}\\
        % \begin{overpic}[height=3.8cm,keepaspectratio]{figs/example_plane_tSNE.pdf}
        %     \put(0,90){t-SNE}
        % \end{overpic}
        \small
        \textbf{(b) t-SNE}\\
        Stress ($\downarrow$): $0.0764$\\
        Trustworthiness ($\uparrow$): $0.9993$
    \end{minipage}
    \caption{Projections of the \textit{plane} dataset with PCA and t-SNE. Common quality metrics such as stress (lower is better) and trustworthiness (higher is better) barely distinguish them, despite significant visual differences.
    %Despite significant visual differences, common quality metrics such as stress (lower is better) and trustworthiness (higher is better) barely distinguish the two projections.
    }
    \label{fig:examples-gaps}
\end{figure}

%
% gap in the literature
%
While widely used by DR practitioners, common quality metrics are not necessarily sensitive to changes that have a high visual impact.
Indeed, recent literature raised awareness about cases where a significant distortion in the visual layout is not properly captured by quality metrics~\cite{Machado_Behrisch_Telea_2025}; this can lead to overconfidence in the analysis of the projection.
An example of such an issue is presented in \cref{fig:examples-gaps}. The \textit{plane} dataset consists of $2$k points randomly distributed on a plane and embedded in 3D space.
Compared to the PCA projection~\cite{jolliffe2005}, which captures the plane perfectly, a local technique such as t-distributed Stochastic Neighbor Embedding (t-SNE)~\cite{vandermaaten08a} (with a perplexity value of $30$) generates spurious patterns, a well-known and expected outcome~\cite{wattenberg2016how}. This distortion has a high visual impact and can mislead visual analysis; however, it is minimal in terms of per-point structural preservation, and common metrics such as scale-normalized stress~\cite{Smelser_Miller_Kobourov_2024} (from now on referred simply as ``stress'') and trustworthiness~\cite{Venna_Kaski_2006}, reported in \cref{fig:examples-gaps}, struggle to capture it.

Machado et al.~\cite{Machado_Behrisch_Telea_2025} argue that standard quality metrics are built on relatively simple approaches with limited scope (e.g., comparing pairwise distances or per-point neighborhoods) and thus cannot capture the richer patterns that humans might see in a projection, such as class clusters and their visual separation.
This raises the need to develop quality metrics that better correlate with the human perception of projections.
Although the problem of assessing the quality of cluster separation in a projection has already been widely studied (see ~\cref{sec:rw-metrics}), less work has been done to measure that of empty regions, which inherently define clusters and other patterns important for visual analytics tasks~\cite{abbas_2019, Giesen_2017}.
Indeed, the Gestalt law of proximity~\cite{WARE2013179} states that groups of points placed close together are perceptually and pre-attentively grouped (in the case of DR, perceived to share a common set of abstract features not directly visualized~\cite{Nonato_Aupetit_2019}).
Since groups and patterns of points are defined by the empty gaps in the projection (or lack thereof)~\cite{ocallaghan_1974,quadri_2021,sadahiro_1997}, it is important to ensure that they are adequately displayed.

In this paper, we address this issue by presenting the Gap Index (GI), a quality metric that quantifies projection distortion from a visual analysis perspective, with an explicit focus on the reliability of visible gaps in the layout.
Contrary to popular quality metrics that compute distortion per-point (and later averaged to quantify the whole layout), the GI focuses on capturing the distortion of the empty areas between points.
This makes the metric operate on a local scale and be sensitive to visually salient distortions, while capturing global distortion with a behavior similar to stress.

To compute the distortions, the GI partitions the 2D projection into triangles and computes their deformation by comparing them with their high-dimensional counterparts. The set of per-triangle deformation measures can be aggregated into a single scalar value or displayed in the scatterplot to visualize distortion by region.
The metric can be efficiently computed in $O(N \log N)$ time, making it fast to run even for large datasets, and does not require explicit coordinates of the original points, as long as a set of triangular pairwise distances is provided.

The core ideas of the GI were previously described by Ros et al.~\cite{ros2025}; in this paper, we revisit and extend them. Additionally, we provide detailed use cases for the GI and comparisons with other popular quality metrics.

Our contributions are the following:
(1)~we identify limitations in the usage of common quality metrics for visual analysis, and we propose a novel perspective for evaluating the quality of DR projections;
(2)~we introduce the GI, a quality metric that measures the distortion in the visual features of a scatterplot; and
(3)~we present a collection of artificial and real datasets used to benchmark the behavior of the GI alongside other popular quality metrics.

An open-source Python implementation of the GI, along with code to reproduce the results presented in this paper, is available online~\cite{ros2026_codeberg}.

%%%%%%%%%%%%%%%%%%%%%%%%%%%%%%%%%%%%%%%%%%%%%
%%%%%%%%%%%%%%%%%%%%%%%%%%%%%%%%%%%%%%%%%%%%%
%%%%%%%%%%%%%%%%%%%%%%%%%%%%%%%%%%%%%%%%%%%%%
\section{Background and Related Work}

In this section, we introduce background concepts and previous work that will help readers follow the rest of the discussion in the paper.

\subsection{Evaluating DR projections} \label{sec:rw-background}

Quality metrics for DR, also known as \textit{distortion measures}, compute how faithful a low-dimensional projection (usually 2D) is as a representation of the original high-dimensional data, following different criteria in terms of which type of high-dimensional structures are aimed to be preserved~\cite{Jeon_Cho_2023}.

The application of quality metrics is twofold. Firstly, they are used as an objective function in DR techniques to find the optimal projection of the high-dimensional data according to that criterion~\cite{mead_1992, Venna_2010, Inaba_Salles_Rauber_2011}; secondly, DR practitioners use them to assess the quality of the final result. In this paper, we focus mainly on the latter.

With the exception of DR methods that provide an explicit mapping to Cartesian coordinates, such as PCA, the absolute positions of points in the 2D layout carry no information; only relative positions with respect to other points are relevant, and scatterplots are typically displayed without orthogonal axes and other frames of reference~\cite{Nonato_Aupetit_2019}. For the same reason, quality metrics are generally invariant to a rigid transformation of the projection.

Another possible transformation of the layout is uniform scaling by a factor $\alpha$. Since users tend to ignore scaling during visual analysis, it is generally desired that quality metrics also be invariant to this transformation (\textit{scale-invariant}, sometimes referred to as \textit{normalized}), although this is not always the case~\cite{Smelser_Miller_Kobourov_2024}.
Scale-invariant metrics tend to be bounded within a certain range (typically $[0,1]$), making them easier to interpret~\cite{Nonato_Aupetit_2019}; the GI is one such metric.

\subsection{Standard quality metrics}~\label{sec:rw-metrics}

Quality metrics generally aim to measure how well the structure of the original data is preserved in the projection. There exists a multitude of metrics, each focusing on different properties of the data structure. A common taxonomy divides them by their target structural granularity, where metrics can be considered \textit{global}, \textit{local}, and \textit{cluster-level}~\cite{Jeon_Cho_2023}.
In the following, we provide relevant examples of popular quality metrics; for a more complete discussion of global and local metrics, we refer the reader to the survey by Nonato and Aupetit~\cite{Nonato_Aupetit_2019}.

\textbf{Global metrics} quantify distortion by measuring the preservation of distances between all points. Stress metrics compute the square difference of pairwise distances between the original and projected points and are the most common type of quality metric in this category. There exist multiple variations of it~\cite{mead_1992}, such as raw stress~\cite{Torgerson_1952}, Kruskal stress~\cite{kruskal1964}, Sammon stress~\cite{sammon1969nonlinear}, and scale-normalized stress~\cite{Smelser_Miller_Kobourov_2024}.
Unless otherwise specified, throughout this paper, we use scale-normalized stress as the preferred type of stress measure.
A popular alternative to stress is correlation-based metrics, such as Shepard goodness~\cite{Espadoto_Martins_Kerren_Hirata_Telea_2021}, which computes the Spearman rank correlation of the Shepard diagram~\cite{Joia_Paulovich_Coimbra_Cuminato_Nonato_2011} (i.e., the correlation between the set of pairwise distances in high- and low-dimensional spaces), or the correlation coefficient~\cite{Geng_Zhan_Zhou_2005}.

On the other hand, \textbf{local metrics} operate in the neighborhood of each point and measure, element-wise, the number of neighbors in the original data which are no longer neighbors in the projection (missing neighbors), and the number of neighbors in the projection which are not actually neighbors in the original data (false neighbors)~\cite{Nonato_Aupetit_2019}.
Trustworthiness and continuity~\cite{Venna_Kaski_2006} are two well-known metrics in this category; they measure the degree to which false and missing neighbors occur for each point. Smooth neighborhood preservation~\cite{Pagliosa_Paulovich_Minghim_Levkowitz_Nonato_2015} measures both the number of neighbors preserved and their change in distance.
Lee and Verleysen~\cite{Lee_Verleysen_2009} review some of the most popular quality metrics in this category.

Finally, \textbf{cluster-level metrics} measure the preservation of cluster structures between the original and projected data; examples are steadiness and cohesiveness~\cite{Jeon_Ko_Jo_Kim_Seo_2022}.
It should be noted that the metrics listed in this section are \textit{unsupervised} since they do not make use of external information, such as class labels.
There exists a broad collection of supervised metrics, especially of cluster-level granularity, that make use of labeled data, such as distance consistency~\cite{Sips_2009}, neighborhood hit~\cite{Paulovich_Nonato_Minghim_Levkowitz_2008}, and GONG~\cite{Aupetit_2016}; however, these are intended mainly to be used in supervised tasks (e.g., projecting data in order to maximize class separation~\cite{Wang_2018}), rather than visual analysis of the data.

%We consider the GI to be in the category of unsupervised local metrics, since only neighboring points are taken into account when computing the deformation (\textit{neighboring} with respect to a specific triangulation, rather than nearest neighbors from a distance function), and it does not make use of any external information.
%However, it differs from other popular local quality metrics by measuring deformations in empty regions of space rather than in neighborhoods around individual points, resulting in greater sensitivity to distortion in visual gaps.
The previous taxonomy is based on metrics that measure distortion from the placement of the points.
The GI operates differently, accounting for distortion in the gaps of the projection; therefore, it cannot be clearly classified into any of the previous categories.
From a distortion-per-point perspective, the GI behaves, on the one hand, as a local metric, since the position of a point only affects the quality values of the triangles incident to it; on the other hand, it captures distortion between points that are arbitrarily far away, as long as they are separated by empty space, which only global metrics measure.

%Liu et al.~\cite{Liu_Wang_Bremer_Pascucci_2014} introduced a quality metric to measure distortion through the change in density around a point. The density is computed with a Gaussian kernel, and a hyperparameter $\sigma$ allows it to be used at different resolutions.
%While this approach indirectly measures the preservation of empty space, it is constrained to the neighborhood of points; thus, large empty areas will not be measured.

The foundational idea for the GI was first proposed by Warnking et al.~\cite{WARNKING20021665}, who introduced the concept of triangulating the 2D projection and computing the compression ratio for each triangle.
Later, Aupetit~\cite{Aupetit_2007} further formalized the method into a measure of local distortion for each triangle; however, this approach was intended primarily as a tool to visualize compressed and stretched regions of a projection.
With the GI, we develop it into a full projection quality metric, including a normalization of the areas to ensure scale invariance, and an aggregation of all local values into a single scalar while taking into account the relative importance of each.

\subsection{What the user sees vs. what the user should see}

The distortions in the projection, with regard to visual analysis, are typically described as having two forms~\cite{Nonato_Aupetit_2019}: data structures in the high-dimensional space that are not observable in the projection, and structures and patterns visible in the projection that do not exist in the original data.
This distinction plays an important role in visual analysis, since it captures two different phenomena that users might want to check separately: ``are there features in the data that are not visible?'', and ``are the visible features real?''.
In the context of neighborhood preservation, this is sometimes expressed with the duality of \textit{missing neighbors} and \textit{false neighbors}, presented in the previous section.

Most quality metrics do not distinguish between the two; instead, they compare the high-dimensional space and its projection simultaneously (e.g., stress compares both sets of pairwise distances).
However, there are metrics that take this distinction into account and come in two dual forms, each capturing one type of deformation. A well-known example is trustworthiness and continuity~\cite{Venna_Kaski_2006}, which measure the preservation of local neighborhoods across spaces. Steadiness and cohesiveness~\cite{Jeon_Ko_Jo_Kim_Seo_2022} follow the same approach at the cluster level.
The GI uses a similar approach but focuses primarily on assessing the projection's veracity.
As such, it captures simultaneously (1)~whether gaps in the layout are also gaps in the original data, and (2)~whether non-gaps in the layout are also non-gaps in the original data.

\subsection{Visualizing distortion in the plot} \label{sec:rw-visualization}

Quality metrics typically produce a single scalar value that quantifies the total distortion of a projection. While this is convenient for numerically comparing projections, it offers limited detail for the visual analysis of 2D projections. However, most metrics allow one to compute the distortion for each individual point. Rather than averaging to quantify the quality of the whole projection, one can visually analyze the local data; this allows users to see which regions of the scatterplot are better preserved and which have experienced greater distortion.
This approach provides more detailed information about the distortion observed in the data, but it also poses challenges for visualization.

The easiest and most common method is to color the points of the projected layout according to a specific scalar quality metric~\cite{mokbel2013visualizing, hermann2009interactive}. However, this approach can suffer from scalability issues, as layouts must ensure that all individual points remain legible and often require space-filling techniques~\cite{Martins_Coimbra_Minghim_Telea_2014} or binning~\cite{orlov_2025}. More advanced approaches use interactivity to enhance the analysis of the visualization~\cite{Martins_Coimbra_Minghim_Telea_2014,Heulot_Aupetit_Fekete_2013}.
Aupetit~\cite{Aupetit_2007} proposed the idea of displaying per-point quality information in the background of a scatterplot by coloring the Voronoi cell of each point according to its quality.
This was further developed in the CheckViz method~\cite{Lespinats_Aupetit_2011}, which uses a 2D color map to encode two types of distortion simultaneously. The authors argue that a colored background provides contextual information about the foreground while avoiding perceptual side effects produced by coloring the points.
However, they acknowledge two shortcomings: first, larger Voronoi cells around isolated points are more visible, creating an area bias; second, this visualization can mislead the user into inferring that the background color quantifies deformation in the empty space itself, whereas it only displays information about the point. 
Notice that both of these issues do not affect our approach, since we color empty areas according to the distortion measured on the empty areas themselves.

% Other possible RW:
% - Reliability map for visualization (Steadiness and Cohesiveness paper). Seems a bit ad-hoc for those metrics, even though they implement it in Zadu
% - Scagnostics (From High Dimensions to Human Insight). They quantify features of the plot (presence of outlying data, clear trends, etc.), but by themselves do not measure quality
% - Decision maps (color background based on NN). Not very related

%%%%%%%%%%%%%%%%%%%%%%%%%%%%%%%%%%%%%%%%%%%%%
%%%%%%%%%%%%%%%%%%%%%%%%%%%%%%%%%%%%%%%%%%%%%
%%%%%%%%%%%%%%%%%%%%%%%%%%%%%%%%%%%%%%%%%%%%%
\section{Methodology} \label{sec:methodology}

In this section, we provide intuition and formalize the Gap Index (GI) as a quality metric, including its computation and visualization overlaid on a DR scatterplot.
In addition to the main proposed approach, we also provide alternatives for each step of the computation, which DR practitioners may find useful for tuning the GI to explore different types of distortion.

To set notation, let $\mathbf{X} = \{ x_i \in \mathbb{R}^q \mid 1 \le i \le N\}$ be a projection of a set of $N$ high-dimensional points $\mathbf{\hat{X}} = \{ \hat{x}_i \in \mathbb{R}^D \mid 1 \le i \le N\}$, with $D > q$ (in the context of visual analysis, $q=2$). A quality metric is a function $\mathcal{Q}(\mathbf{X}, \mathbf{\hat{X}}) \rightarrow \mathbb{R}$ that quantifies the degree of structural preservation of $\mathbf{X}$ with respect to $\mathbf{\hat{X}}$. 

Among the various structural distortions introduced by a DR technique, the GI focuses on measuring the distortion of layout gaps or empty spaces, in terms of stretching or compression, relative to the complete layout. It does so by comparing the proportion of space that the regions in $\mathbf{X}$ use, compared to the same regions in $\mathbf{\hat{X}}$.

\subsection{Overview of the GI}

%\begin{figure}
%    \centering
%    \includegraphics[width=0.55\linewidth]{figs/pipeline.pdf}
%    \caption{Overview of the computation of the GI.}
%    \label{fig:pipeline}
%\end{figure}

The computation of the GI given $\mathbf{X}$ and $\mathbf{\hat{X}}$ follows a modular structure, as shown in \cref{fig:teaser}:
first, a \textit{triangulation} of the points is constructed on the projection, partitioning $\mathbf{X}$ into a set of triangles $\mathcal{T}(\mathbf{X}) = \{ t_1, t_2, \dots, t_m \}$. Subsequently, we compute the \textit{deformation} of each triangle $t_i$, by relating it to the triangle $\hat{t}_i$ formed by the same three points in $\mathbf{\hat{X}}$; finally, the user can choose to perform an \textit{aggregation} step, to obtain a single scalar value $\mathcal{Q}(\mathbf{X}, \mathbf{\hat{X}})$ quantifying the overall distortion of the projection, or to \textit{visualize} the distorted regions on top of the scatterplot.
These steps are detailed in the following subsections.

We make no assumptions about the projection technique, except that it produces a 2D configuration of data points ($\mathbf{X}$). The GI is intended primarily to be used with two-dimensional projections, which is the most common scenario in the use of DR for visual analysis of high-dimensional data; however, in \cref{sec:conclusion} we discuss its possible use to evaluate projections of higher dimensionality.
Additionally, the GI only requires pairwise triangular distances between the original points $\mathbf{\hat{X}}$, rather than explicit coordinates for each point.

\subsection{Triangulation} \label{sec:met-triangulation}

The GI, rather than focusing on individual points, operates on the empty spaces, or gaps, in the 2D projection. Thus, the first step is to subdivide the visual space into smaller regions for which to compute the distortion.
We propose to decompose it into triangles defined by the projected points, motivated by multiple factors:
(1)~a triangle is a simple primitive, which simplifies computations;
(2)~any triangle defined by three points in $\mathbf{X}$ will have a corresponding triangle in the original space, possibly with zero area, defined by the corresponding points in $\mathbf{\hat{X}}$ (this is not the case with primitives of higher order, such as quads, since unless the points are planar in high-dimensional space, their areas are ill-defined);
(3)~trigonometric operations allow one to compute properties of a triangle defined only by the lengths of its edges, which is needed if $\mathbf{\hat{X}}$ is given simply as a set of distances, rather than explicit high-dimensional Euclidean coordinates;
finally, (4)~there exist simple and well-known methods to decompose the 2D space into a set of triangles.

Our method of choice for this step is the Delaunay triangulation, a widely used and well-studied triangulation technique~\cite{fortune2017voronoi}.
It decomposes the 2D space into triangles with the layout points as vertices. These triangles are non-overlapping and empty, and they cover the full convex hull of the projected points, which we consider the total visual space of a projection.
Moreover, it maximizes the minimum angle among all triangles, avoiding thin triangles that can cause numerical instability~\cite{De_Berg_2008}.

The Delaunay triangulation also offers computational advantages, since efficient algorithms exist to compute it from a set of 2D points in $O(N \log N)$ time~\cite{De_Berg_2008}, and the number of triangles generated is on the order of $O(N)$~\cite{Seidel_1995}. Since the complexity of the entire GI is dominated by the triangulation, this makes the GI fast to compute as a metric.
Finally, given a set of points, this approach is nonparametric and has a unique solution, assuming that degenerate cases such as cocircular points are resolved deterministically, making the GI also a deterministic metric.

\subsection{Triangle deformation}

The previous step produces a set of triangles $\mathcal{T}(\mathbf{X}) = \{ t_1, t_2, \dots, t_m \} $. Each triangle $t_i \in \mathcal{T}(\mathbf{X})$ is defined by three vertices, which are points in $\mathbf{X}$. We can then consider the corresponding triangle $\hat{t}_i$ in the original space formed by the same three points in $\mathbf{\hat{X}}$ and measure its deformation in the projection.
Our suggested approach is to compare the relative areas of each triangle.
Let $A(t_i)$ denote the area of triangle $i$ in $\mathbf{X}$, and $A(\hat{t}_i)$ the area of the corresponding triangle in $\mathbf{\hat{X}}$. Let $A'(t_i)$ and $A'(\hat{t}_i)$ be the relative areas of each triangle:

\begin{equation}
    A'(t_i) = \frac{A(t_i)}{\sum_j A(t_j)} , \quad
    A'(\hat{t}_i) = \frac{A(\hat{t}_i)}{\sum_j A(\hat{t}_j)}.
\end{equation}

Relative areas compare the proportion of the total space taken by each triangle, which is easier to interpret than their raw values. This also ensures that the metric is scale-invariant, which is a desirable property for metrics intended to capture visual distortions, as motivated in \cref{sec:rw-background} and discussed in \cref{sec:discussion}.
We then define the \emph{deformation} of a triangle $t_i$ as

\begin{equation} \label{eq:deformation}
    D(t_i, \hat{t}_i) = \frac{A'(t_i) - A'(\hat{t}_i)}{max(A'(t_i),A'(\hat{t}_i))},
\end{equation}

where the denominator ensures that the value is in the range $[-1,1]$, with positive values of $D(t_i, \hat{t}_i)$ meaning that the relative size of triangle $t_i$ is bigger than in the original space, indicating a stretching in the visual space. Conversely, negative values of $D(t_i, \hat{t}_i)$ indicate that the region has been compressed in the projection.
This formulation is similar to the one proposed by Aupetit~\cite{Aupetit_2007}, the main differences being the use of relative rather than raw areas, and that we compute a single deformation value instead of two separate ones for compression and stretch.

Note that even if there are no explicit Euclidean coordinates for high-dimensional points, one can compute the area of the triangle only from the distances between points using Heron's formula:

\begin{equation} \label{eq:heron}
    A(a,b,c) = \sqrt{s \cdot (s-a) \cdot (s-b) \cdot (s-c)}
\end{equation}
where $a$, $b$, and $c$ are the edge lengths of the triangle (i.e., the distances between the three high-dimensional points) and $s = (a+b+c)/2$. The distances are assumed to fulfill the triangular inequality (otherwise $A(a,b,c)$ will result in a complex value); this assumption is reasonable, since a set of non-triangular distances cannot be meaningfully projected in 2D space seeking to preserve the distances or any similarity (e.g., the t-SNE probabilities) derived from it.
Since extremely thin triangles can pose precision problems during the computation of the area, we recommend clipping the result beyond a tolerance value (e.g., $10^{-6}$) and considering the area as $0$ (if $A(t_i) = A(\hat{t}_i) = 0$, then $D(t_i, \hat{t}_i) = 0$ by convention); however, it should be noted that there exist alternative formulations for \cref{eq:heron} that are numerically more stable to compute with floating-point precision~\cite{kahan1983mathematics}.

Computing the area of $\hat{t}_i$ through distances, rather than high-dimensional Euclidean coordinates, has the advantage of allowing the use of different distance functions, reflected in the values of $a$, $b$, and $c$.
If one uses a DR technique that optimizes for the preservation of non-Euclidean distances (e.g., Isomap~\cite{Tenenbaum_Silva_Langford_2000}, which uses geodesic distances in a nearest-neighbors graph), it is advisable to use that function also in the computation of $A(\hat{t}_i)$, as it will make the metric more coherent with the projection~\cite{Tsai_2012}.
While in this paper we only consider Euclidean distances for the computation of $A(\hat{t}_i)$, users should be aware of its limitations~\cite{aggarwal2001}, which affect all quality metrics that rely on distances (such as stress, trustworthiness and continuity, or cluster-based metrics).
Particularly for the GI, this can result in all triangles in the original space of high-dimensionality having a similar area, and the metric correlating with the density of the points in 2D. In such cases, alternatives to the distance metric that are more meaningful in a space of high-dimensionality, such as the $L_1$ distance, could be preferred.

\subsection{Aggregation}

To compute a single scalar metric that quantifies the overall distortion of the projection, we aggregate all triangle deformation values. Since not all triangles in $\mathcal{T}(\mathbf{X})$ are equally relevant to the global result, we propose using a weighted average of their values. A triangle is considered relevant if it has a large relative area in $\mathbf{X}$, in $\mathbf{\hat{X}}$, or both, since those are the ones that encode the largest and most salient gaps, whether in the projection or in the original space; thus, we weight the deformation of a triangle $D(t_i, \hat{t}_i)$ by $w_i = \max(A'(t_i), A'(\hat{t}_i))$.
This gives the final GI value of a projection $\mathbf{X}$ of a set of high-dimensional points $\mathbf{\hat{X}}$:

\begin{equation} \label{eq:aggregation}
    GI(\mathbf{X}, \mathbf{\hat{X}}) = \frac{1}{\sum_{t_i \in \mathcal{T}(\mathbf{X})} w_i} \sum_{t_i \in \mathcal{T}(\mathbf{X})} w_i \left| D(t_i,\hat{t}_i) \right| ,
\end{equation}
where $\mathcal{T}(\mathbf{X})$ is the set of triangles from the triangulation of $\mathbf{X}$, and $\hat{t}_i$ is the corresponding high-dimensional triangle to $t_i$.
$D(t_i,\hat{t}_i) \in [-1,1]$ is the triangle deformation value as defined in \cref{eq:deformation}; notice the absolute sign, since here we only care about the magnitude of the deformation, and opposite deformations should not cancel each other.

The weights $w_i = \max(A'(t_i), A'(\hat{t}_i))$ ensure that larger triangles have a greater influence on the final value.
Indeed, a large triangle in $\mathbf{X}$ represents a salient gap in the projection, which should have an important contribution to the total value (positive if the gap exists in $\mathbf{\hat{X}}$, and negative if it does not); likewise, large deformation values for large triangles in $\mathbf{\hat{X}}$ correspond to important gaps in the original space, and the total quality should be heavily penalized if they are not visible in the 2D layout.
$w_i$ is only small when both $A'(t_i)$ and $A'(\hat{t}_i)$ are small; these are triangles that have little visual relevance.

Other choices of $w_i$ are possible and serve to capture different types of distortion.
One could consider using $w_i = A'(t_i)$, which focuses only on capturing whether the gaps in $\mathbf{X}$ exist in $\mathbf{\hat{X}}$; conversely, $w_i = A'(\hat{t}_i)$ will capture whether the gaps in $\mathbf{\hat{X}}$ are well represented in $\mathbf{X}$.

\subsection{Visualizing the GI} \label{sec:met-visualization}

As mentioned in \cref{sec:rw-visualization}, visualizing the distortion on top of a scatterplot is useful for conveying local distortions, allowing users to see patterns and region-specific distortions that would be obscured when aggregated into a single scalar, as in \cref{eq:aggregation}.

Contrary to previous common quality metrics, which compute per-point distortion and rely on space-filling techniques to color the background of the scatterplot, the GI measures the distortion of the empty visual space; thus, the values $D(t_i, \hat{t}_i)$ can be naturally displayed and interpreted as background.
Since the Delaunay triangulation will cover the entire convex hull of the 2D layout without any overlap, the most natural method of displaying the GI is to color the triangles according to their distortion value $D(t_i, \hat{t}_i)$.

\begin{figure}
    \centering
    \begin{subfigure}[c]{0.43\columnwidth}
        \vspace{0.25cm}
        
        \includegraphics[height=3.5cm,keepaspectratio,alt={Visualization of the GI on the PCA projection of the COIL20 dataset.}]{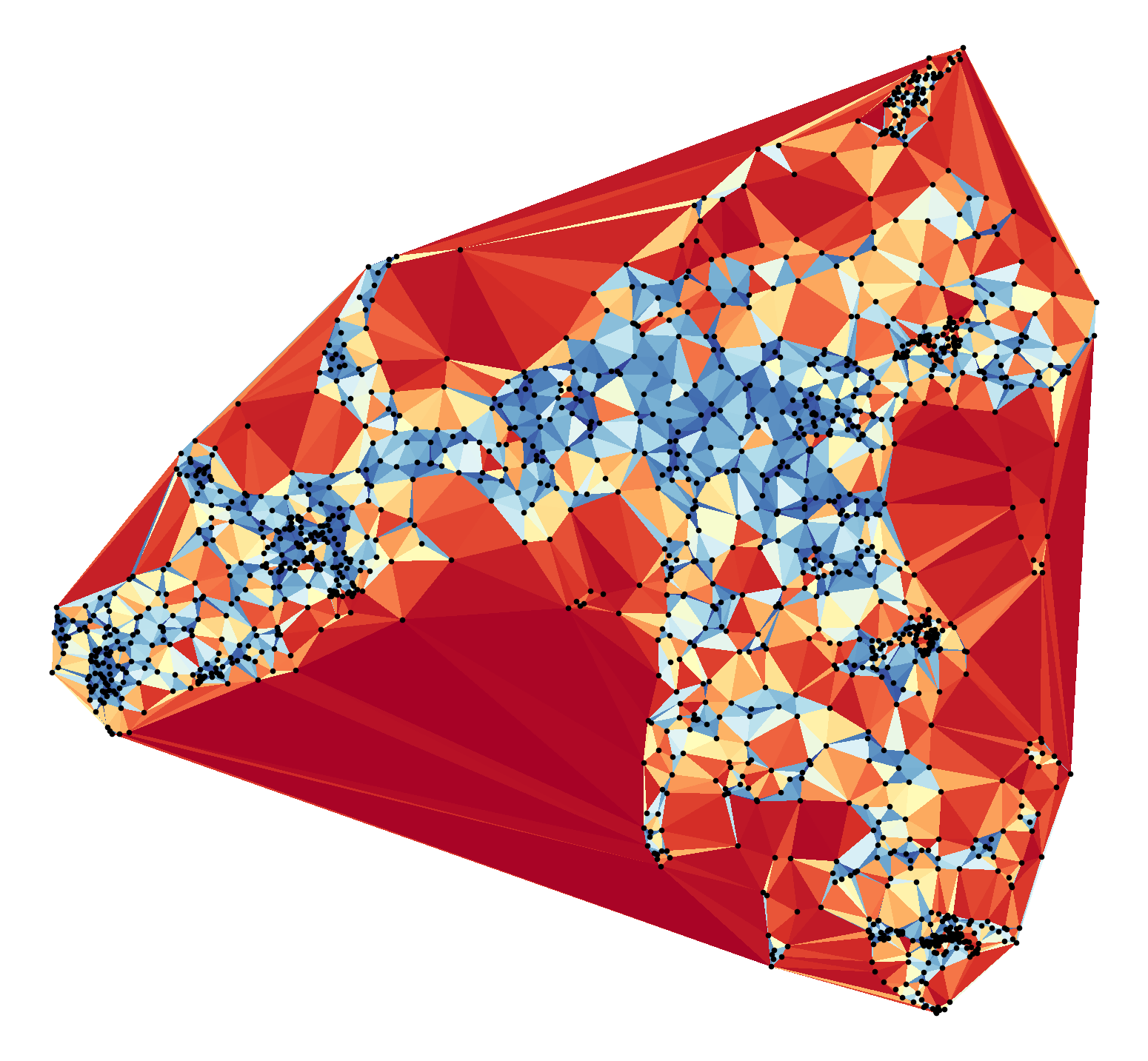}

        \vspace{0.25cm}
        \caption{\textit{COIL20} dataset -- PCA projection}
    \end{subfigure}
    \begin{subfigure}[c]{0.43\columnwidth}
        \includegraphics[height=4cm,keepaspectratio,alt={Visualization of the GI on the t-SNE projection of the Fashion MNIST dataset.}]{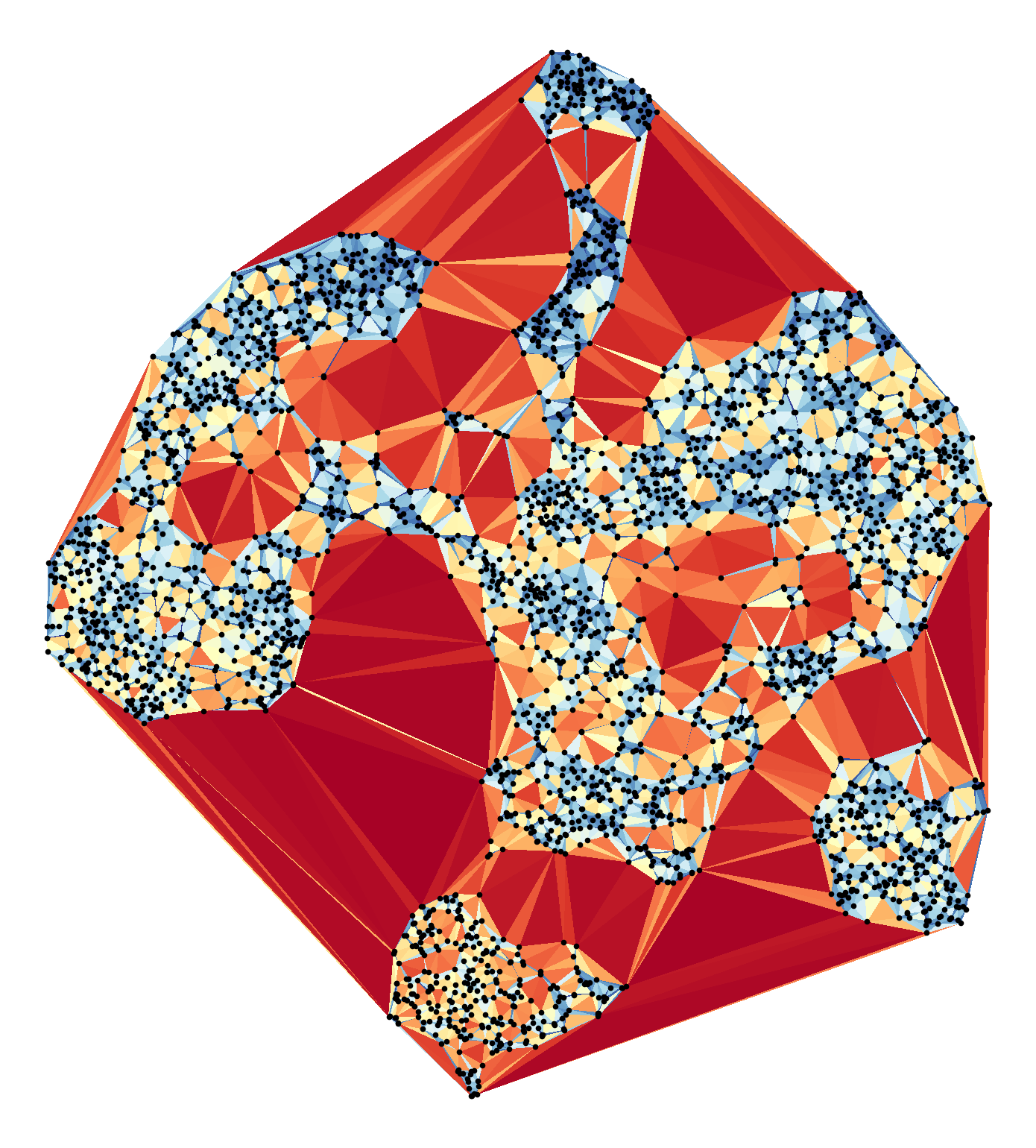}
        \caption{\textit{Fashion MNIST} dataset -- t-SNE projection}
    \end{subfigure}
    \begin{subfigure}[c]{0.08\columnwidth}
        \includegraphics[height=4cm,keepaspectratio,alt={Colorbar blue-yellow-red.}]{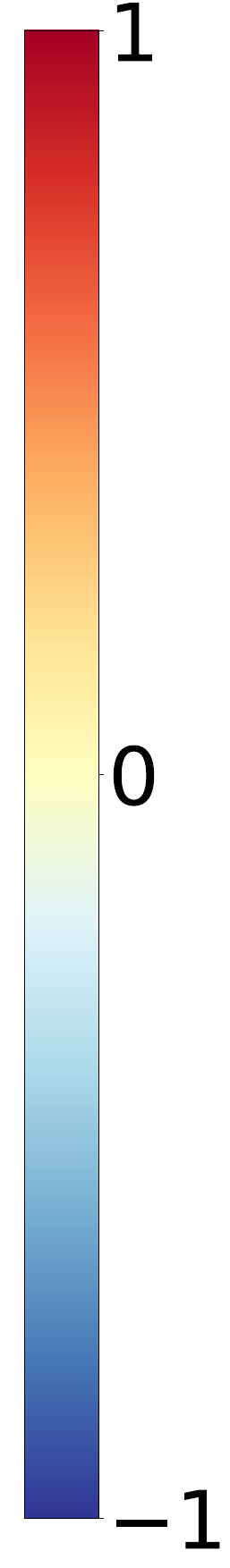}
        \caption*{}
    \end{subfigure}
    \caption{Visualization of the GI on two projections. Red indicates a stretch of the space; blue, a compression; and yellow, no distortion.}
    \label{fig:visualization}
\end{figure}

This is shown in \cref{fig:visualization}.
We use a divergent red-yellow-blue color scheme to encode the two opposite distortions: red indicates a stretch (the triangle has a larger relative area than in $\mathbf{\hat{X}}$), and blue indicates a compression (the triangle has a smaller relative area than in $\mathbf{\hat{X}}$).
Yellow indicates no distortion; we use yellow rather than white to ensure that it remains distinguishable against a white background.

% \begin{figure}
%     \centering
%     \begin{subfigure}[c]{0.43\columnwidth}
%         \includegraphics[height=3.5cm,keepaspectratio]{figs/blur-coil20-PCA.png}
%     \end{subfigure}
%     \begin{subfigure}[c]{0.43\columnwidth}
%         \includegraphics[height=4cm,keepaspectratio]{figs/blur-fashion-tSNE.png}
%     \end{subfigure}
%     \begin{subfigure}[c]{0.08\columnwidth}
%         \includegraphics[height=4cm,keepaspectratio]{figs/colorbar.pdf}
%     \end{subfigure}
%     % \raisebox{-0.5\height}{\includegraphics[height=3.5cm,keepaspectratio]{figs/blur-coil20-PCA.png}}
%     % \raisebox{-0.5\height}{\includegraphics[height=4cm,keepaspectratio]{figs/blur-fashion-tSNE.png}}
%     % \hspace{1em}
%     % \raisebox{-0.5\height}{\includegraphics[height=4cm,keepaspectratio]{figs/colorbar.pdf}}
%     \caption{Visualization of the GI distortion with a smooth background for the same projections from Figure~\ref{fig:visualization}. The blurred background makes the triangulation less visible and allows users to focus on the projected points and the main distortion patterns at a coarser granularity.}
%     \label{fig:blur}
% \end{figure}

\begin{figure}
    \centering
    \includegraphics[width=0.24\linewidth,alt={Visualization of the GI on the t-SNE projection of the Fashion MNIST dataset with no blur.}]{figs/color-fashion-tSNE.png}
    \includegraphics[width=0.24\linewidth,alt={Visualization of the GI on the t-SNE projection of the Fashion MNIST dataset with little blur.}]{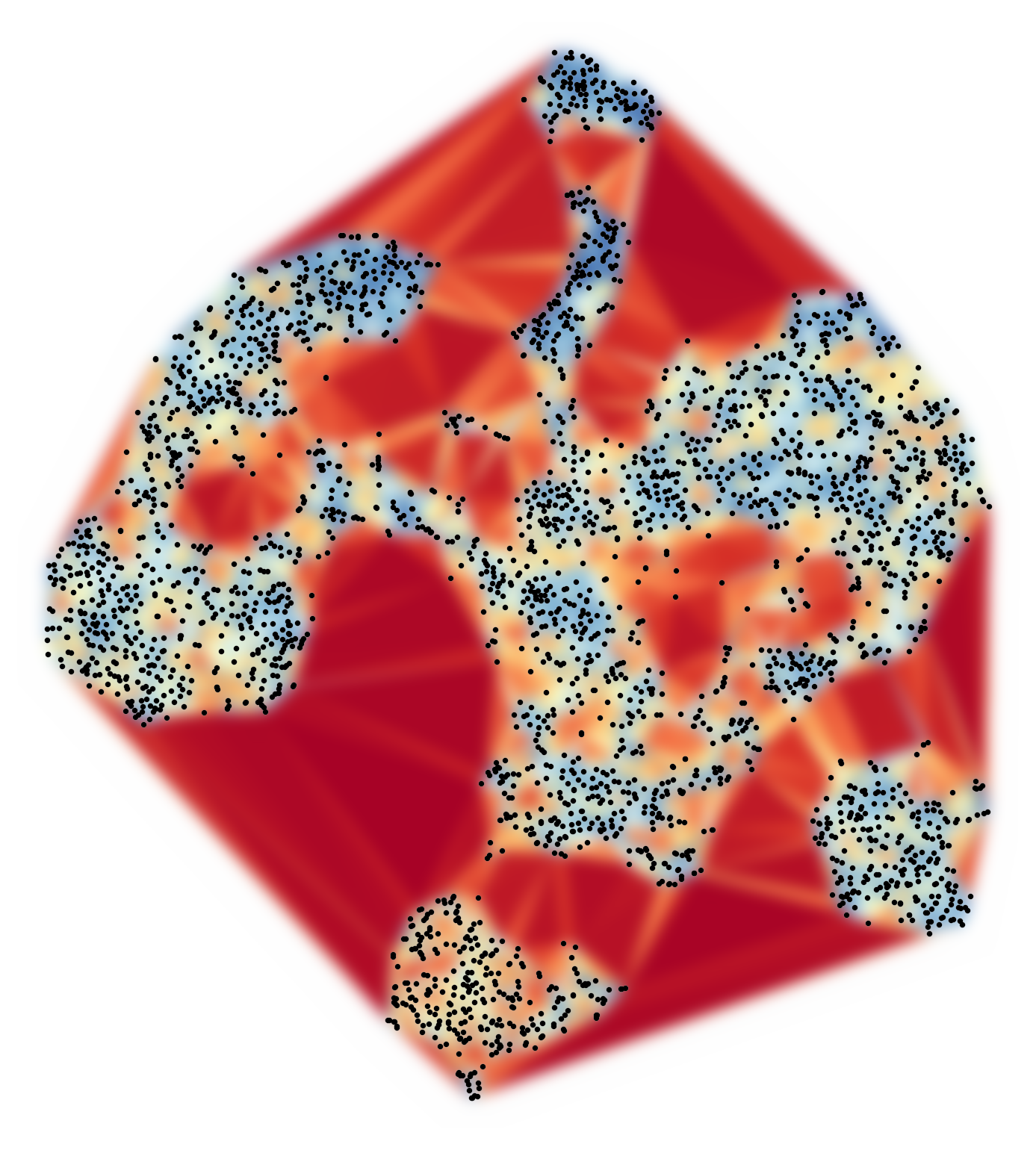}
    \includegraphics[width=0.24\linewidth,alt={Visualization of the GI on the t-SNE projection of the Fashion MNIST dataset with medium blur.}]{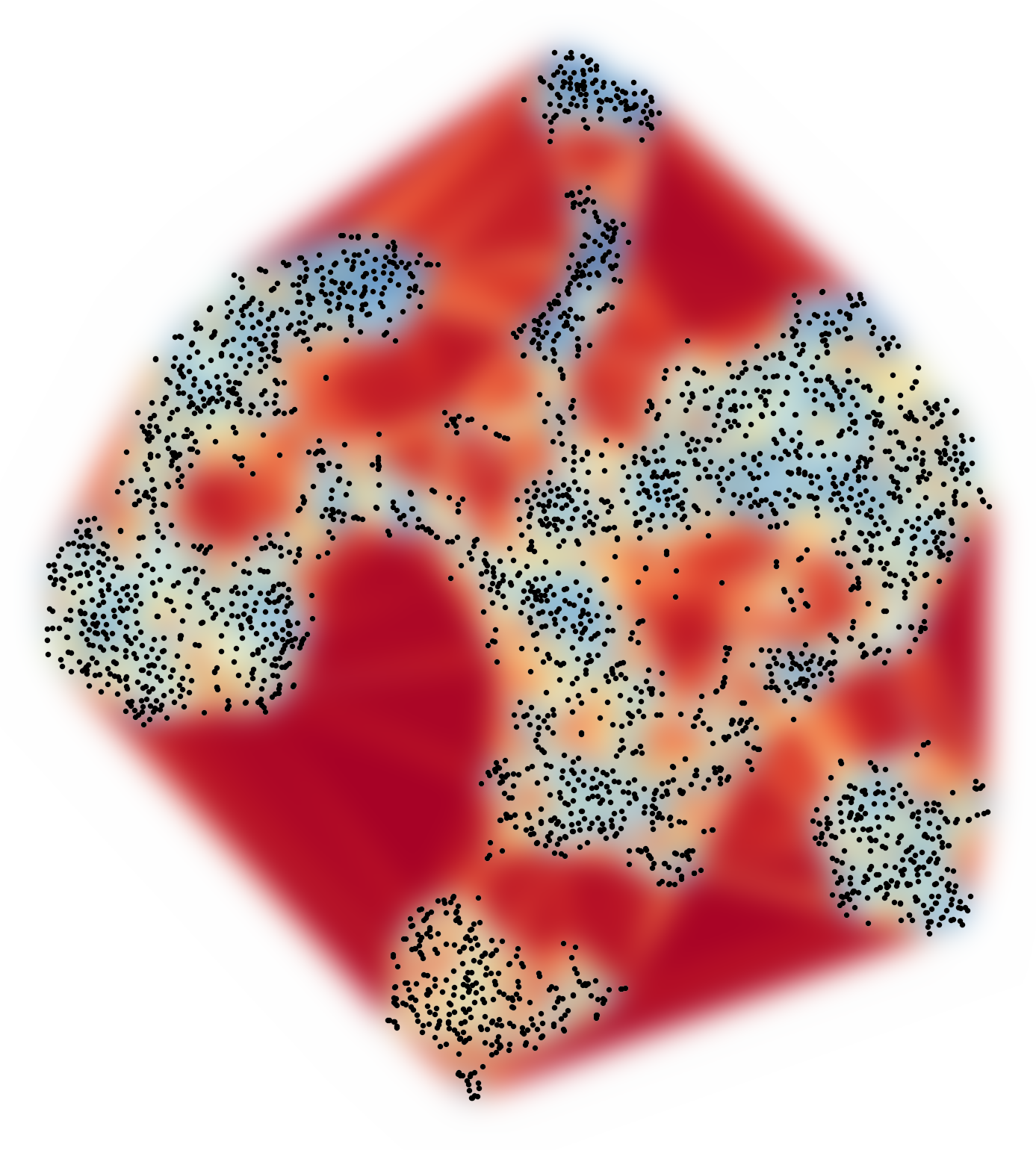}
    \includegraphics[width=0.24\linewidth,alt={Visualization of the GI on the t-SNE projection of the Fashion MNIST dataset with high blur.}]{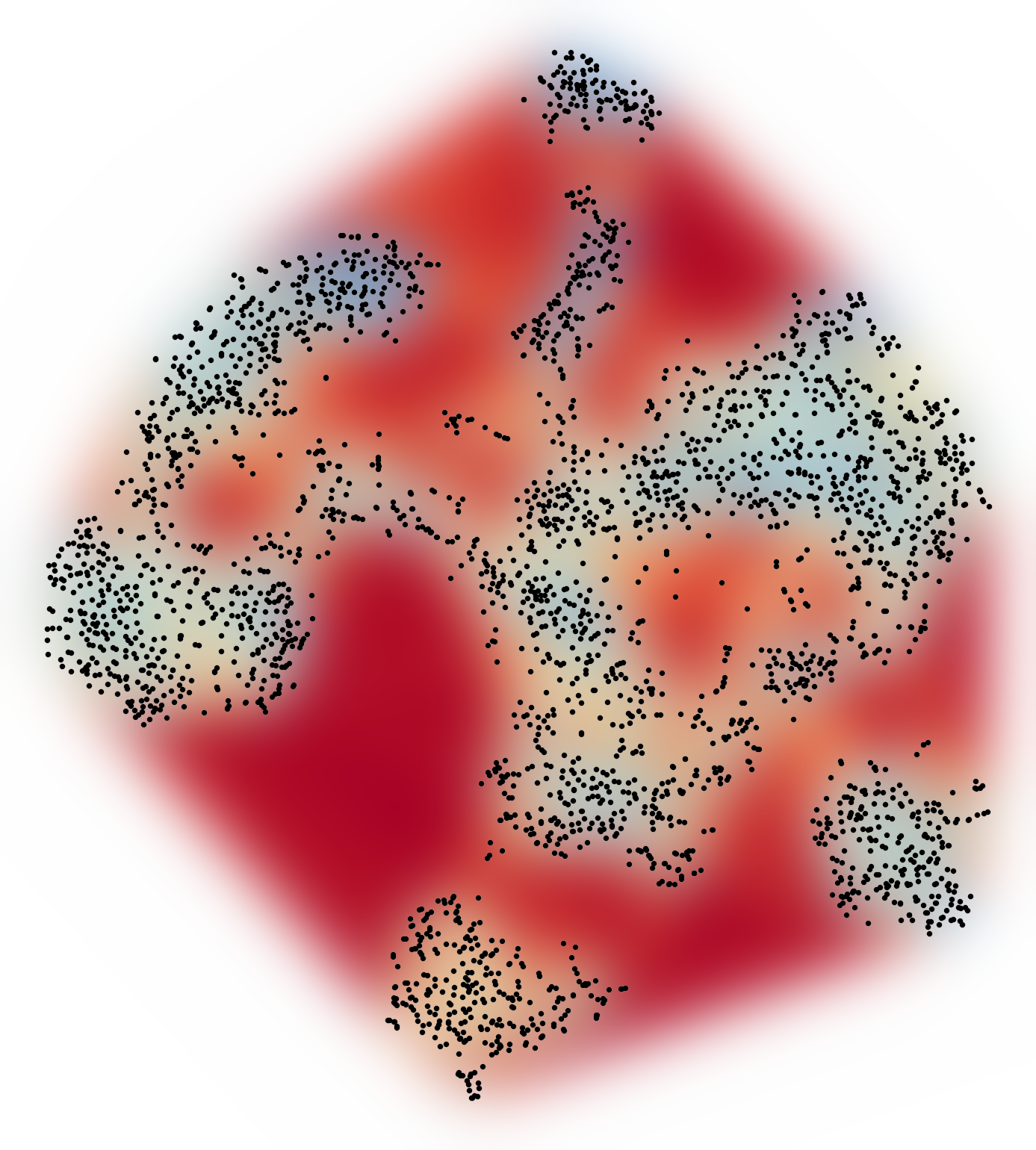}
    \caption{Visualization of the GI on a t-SNE projection of the \textit{fashion MNIST} dataset with different levels of background smoothing. From left to right, the blur levels are: $\sigma=0$, $\sigma=10$, $\sigma=20$, $\sigma=40$.}
    \label{fig:blur}
\end{figure}

Looking at \cref{fig:visualization}, a shortcoming of this approach becomes evident: the triangles distract from the visualization of the overall distortion, despite them bearing no connection to the data beyond being a particular subdivision of the 2D space.
Even though edge lines are not drawn, the hard boundaries between adjacent areas make the underlying triangulation easily visible; moreover, thin, elongated triangles, which generally have small area and thus contribute little to the final value, can be visually salient, complicating the visual analysis of the distortion.
While directly observing the triangles can be useful for a more detailed understanding of the individual deformation values, users seeking a general assessment of the projection distortion might prefer a less detailed view.
To overcome this issue, we propose to smooth the colored background, as shown in \cref{fig:blur}. This ensures that the main distortion patterns remain visible, while the boundaries between individual triangles are softened.
In our example, we consider a uniform Gaussian blur with a user-defined standard deviation $\sigma$ to control the degree of smoothing. 
Higher levels of blur provide a coarser picture of the local distortion, while lower levels are useful for showing more specific details, as illustrated in \cref{fig:blur}; the level of blur is left to the user.
%Higher values of $\sigma$ allow the user to focus on more general distortion patterns by hiding the finer detail. This is further discussed in Section~\ref{sec:discussion}.

\section{Results} \label{sec:results}

In this section, we report examples of using the GI across different datasets, compare its behavior to other popular quality metrics, and test its properties under varying conditions.
Firstly, we show the use of the GI in two controlled DR scenarios to illustrate how to interpret the results of the metric.
We then quantitatively compare it to other popular quality metrics and qualitatively compare it with alternatives to visualize local distortion.
Then, we test the scalability of the GI (again, comparing it to other metrics) and the stability of the Delaunay triangulation.
Finally, we explore a practical use case in the DR literature by evaluating a projection of protein data with the GI.

We compare the GI to other popular quality metrics from different categories:
scale-normalized stress~\cite{Smelser_Miller_Kobourov_2024} (referred to here simply as ``stress''), trustworthiness and continuity~\cite{Venna_Kaski_2006}, and steadiness and cohesiveness~\cite{Jeon_Ko_Jo_Kim_Seo_2022}, which are global, local, and cluster-level metrics, respectively.
We include only the results we consider most relevant, but we encourage interested readers to consult the supplementary material for a more extensive collection of results, including plots of all projections.
As mentioned in the introduction, the code for the GI and scripts to reproduce these results are publicly available online~\cite{ros2026_codeberg}. For the other quality metrics, as well as the CheckViz visualization technique, we use the implementation from ZADU~\cite{Jeon_Cho_2023}; the only exception is during the scalability test (\cref{sec:results-scalability}), where we used an adapted implementation of stress and trustworthiness from Espadoto et al.~\cite{Espadoto_Martins_Kerren_Hirata_Telea_2021}, since we found it to be faster and less resource-intensive.

\subsection{Interpreting the GI}\label{sec:interpreting}

\begin{figure}[thbp]
    \centering
    \begin{subfigure}[c]{0.48\columnwidth}
        \includegraphics[height=4cm,keepaspectratio,alt={Visualization of the GI on the PCA projection of the Cube dataset.}]{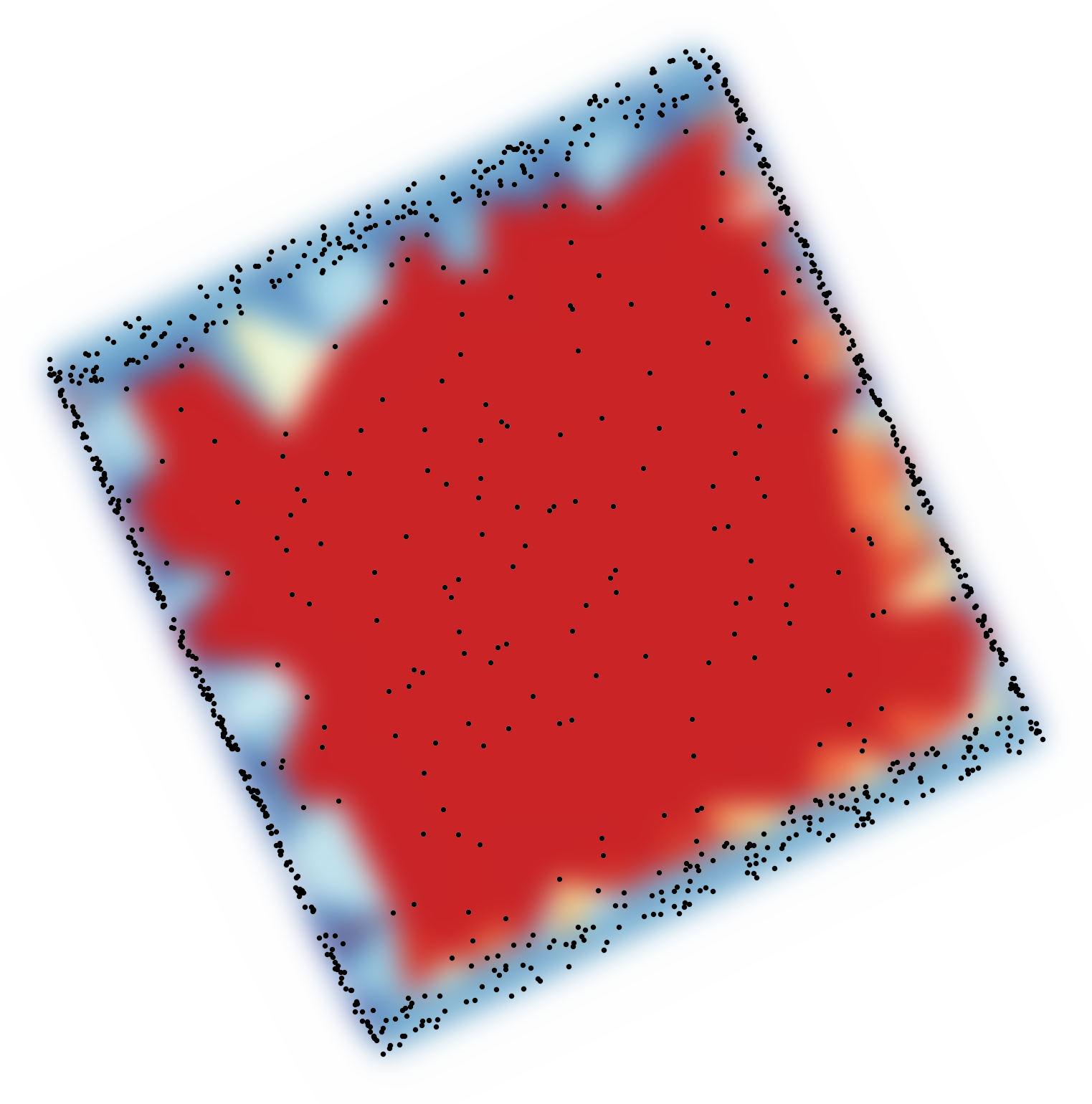}
        \caption{\textit{Cube} dataset -- PCA projection\\GI value: $0.755$}
        \label{fig:results-interpreting-cube}
    \end{subfigure}
    \hspace{0.2em}
    \begin{subfigure}[c]{0.35\columnwidth}
        \includegraphics[height=4cm,keepaspectratio,alt={Visualization of the GI on the t-SNE projection of the Clusters dataset.}]{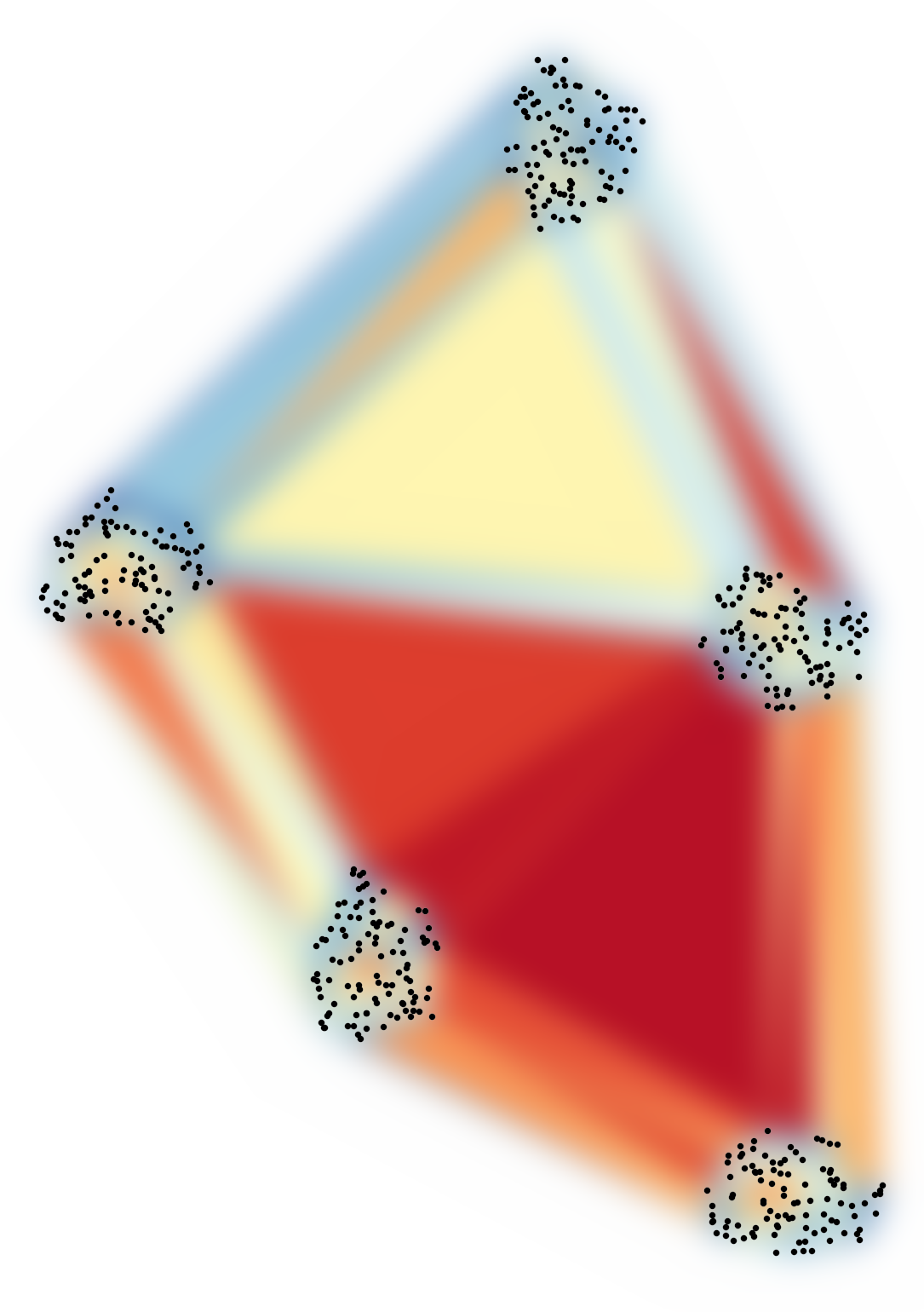}
        \caption{\centering \textit{Clusters} dataset -- t-SNE projection\\GI value: $0.482$}
        \label{fig:results-interpreting-clusters}
    \end{subfigure}
    \hspace{1em}
    \begin{subfigure}[c]{0.08\columnwidth}
        \includegraphics[height=4cm,keepaspectratio,alt={Colorbar blue-yellow-red.}]{figs/colorbar.pdf}
        \caption*{}
    \end{subfigure}
    \caption{Visualization of the GI on projections of two different datasets: (a) an empty 3D cube with a missing face; (b) five equally-sized clusters randomly positioned in 3D space.}
    \label{fig:results-interpreting}
\end{figure}

We use two DR projections of synthetic 3D datasets to showcase the use of the GI and familiarize the reader with the interpretation of the results. The projections, colored according to the GI, are shown in \cref{fig:results-interpreting}.
%Through these controlled three-dimensional cases, for which we can easily comprehend the overall distortion, the interpretation of the GI becomes more intuitive.
The \textit{cube} dataset consists of a set of $1$k points randomly distributed along $5$ sides of a 3D cube. \Cref{fig:results-interpreting-cube} shows its PCA projection, where the aggregated GI has a final value of $0.755$, indicating a highly distorted projection; upon visual inspection, this is evident.
Four of its filled sides have been collapsed into a square with a high density of points on the outline, while the interior of the square is filled with the points of the remaining non-empty face.
Although the points in this fifth face are well positioned (since rigid projections, such as PCA, will never ``stretch'' the data), the empty space between them is disproportionately high compared to the rest of the dataset, spanning almost the entire visual space, whereas in the original 3D manifold it is only a fifth of the surface area; thus, this is shown as a stretched area (red) and heavily penalizes the value of the GI.
The four remaining faces are collapsed into lines, giving the impression of dense areas.

\Cref{fig:results-interpreting-clusters} shows a t-SNE projection of the \textit{clusters} dataset, which consists of $5$ clusters randomly positioned in 3D space; clusters consist of $100$ points each, normally distributed, and with the same variance.
As expected, t-SNE correctly separates the clusters, but, being a local technique, the space between clusters is distorted~\cite{wattenberg2016how}; the GI has a value of $0.482$, indicating substantial distortion. Visualizing the distortion on top of the projection helps explain the reason for this value and infer information about the original 3D structure.
We can see that the space between the top and leftmost clusters has been compressed, indicating they were originally farther apart; the opposite occurs for the other three clusters, where the space between them has increased, indicating they were originally closer.
We invite the reader to check the original 3D data in the supplementary material and verify that this is indeed the case.

\subsection{GI to quantify the overall distortion}\label{sec:results-overall}

In the previous paragraphs, alongside the visualization of the GI, we reported the aggregated value of the GI.
Its main use is to compare different projections of the same dataset; in this case, it helps assess which projection is better and, to some extent, by how much.
However, this single value is of too low granularity for users interested in analyzing the distortions in detail, who should prefer to visualize the metric on top of a scatterplot of the projection.

\begin{figure}[thbp]
    \centering
    \begin{subfigure}{0.43\columnwidth}
        \centering
        \includegraphics[height=3.8cm,keepaspectratio,alt={Visualization of the GI on the PCA projection of the Plane dataset. No distortion is visible.}]{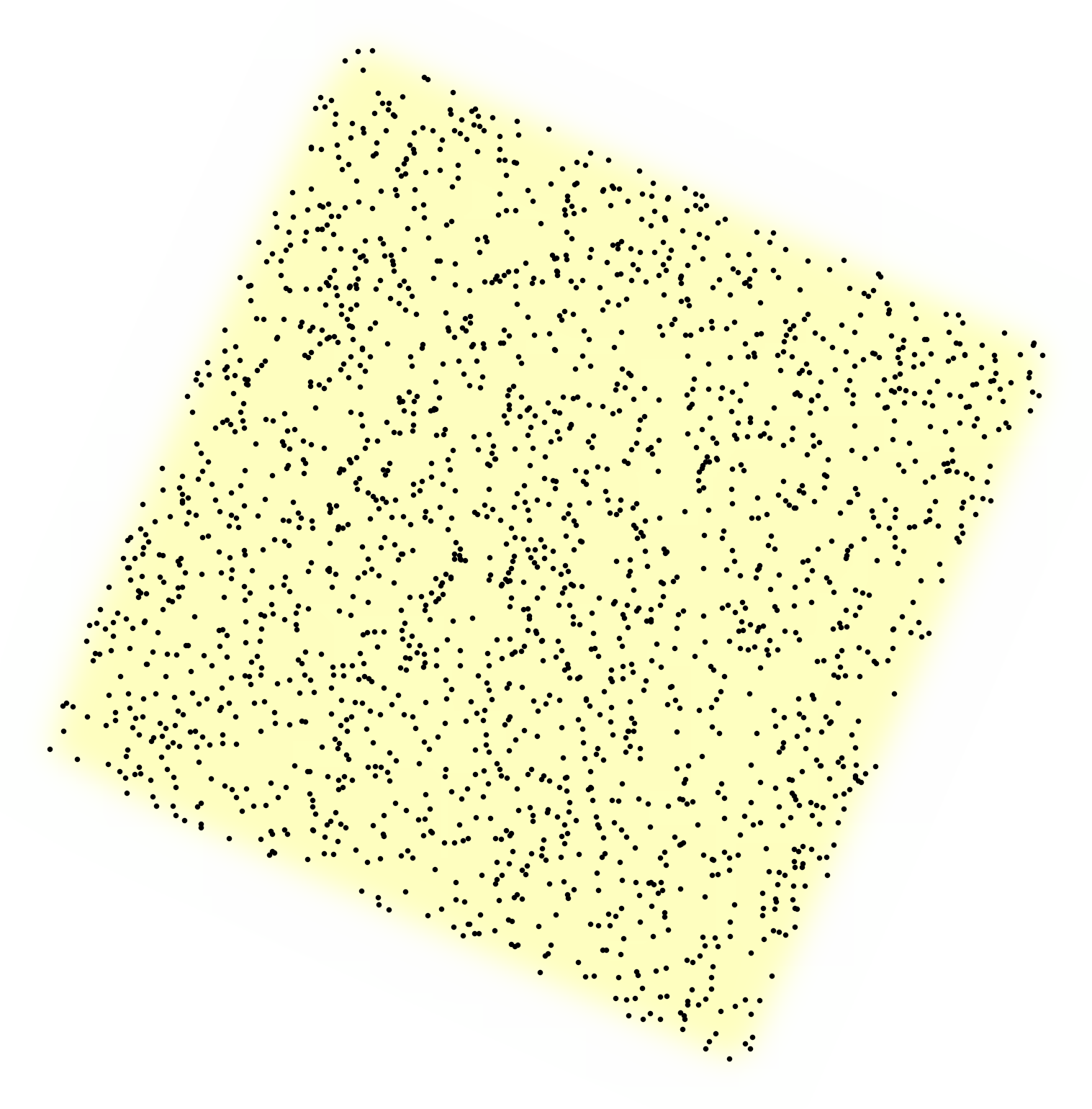}
        \small
        \textbf{(a) PCA}\\
        Stress ($\downarrow$): 0\\
        Trustworthiness ($\uparrow$): 1\\
        \textbf{Gap Index ($\downarrow$): 0}
    \end{subfigure}%
    \hfill
    \begin{subfigure}{0.43\columnwidth}
        \centering
        \includegraphics[height=3.8cm,keepaspectratio,alt={Visualization of the GI on the PCA projection of the Plane dataset. Large empty areas are colored red; dense clusters are colored blue.}]{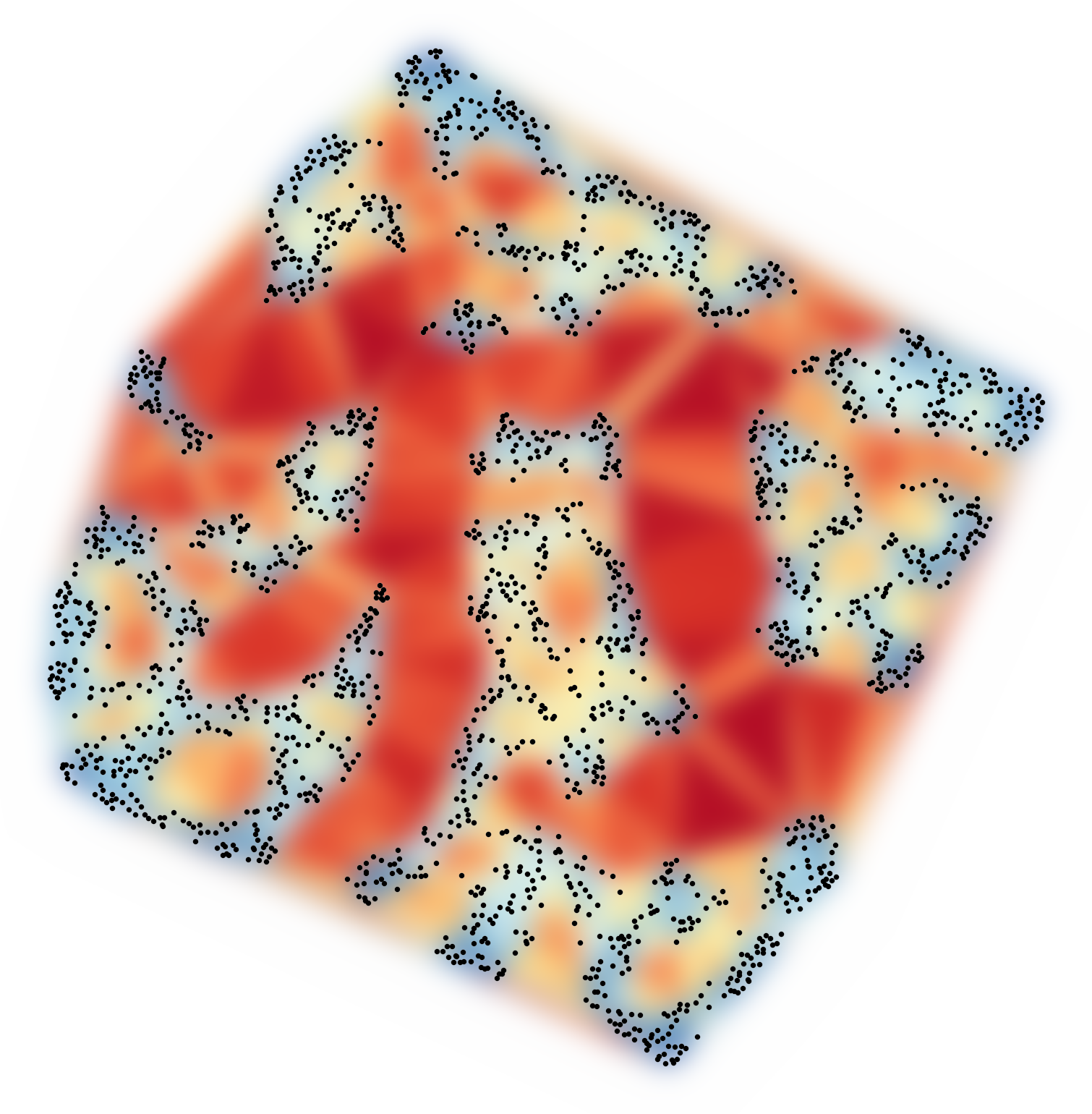}
        \small
        \textbf{(b) t-SNE}\\
        Stress ($\downarrow$): 0.0764\\
        Trustworthiness ($\uparrow$): 0.9993\\
        \textbf{Gap Index ($\downarrow$): 0.5118}
    \end{subfigure}
    \hspace{0.5em}
    \begin{subfigure}{0.1\columnwidth}
        \includegraphics[height=3.8cm,keepaspectratio,alt={Colorbar blue-yellow-red.}]{figs/colorbar.pdf}
        \vspace{4em}
    \end{subfigure}
    \caption{The two projections from \cref{fig:examples-gaps} colored according to the GI. The PCA projection preserves all the gaps perfectly and thus has no distortion. The gaps on the tSNE projection are identified as a stretching of the space (red), while the clusters correspond to a compression (blue).}
    \label{fig:results-plane}
\end{figure}

An example is the \textit{plane} dataset, which has already been introduced in \cref{sec:introduction}, consisting of a set of $2$k points placed randomly in a plane in 3D space (i.e., a uniform distribution in $[0,1]^2$ placed in 3D space with a random rotation). Scatterplots of two projections of it are shown in \cref{fig:examples-gaps}. In \cref{fig:results-plane}, we show the visualization of the GI for the same two projections.
The PCA projection captures all pairwise distances between points, yielding optimal values for all quality metrics. However, the t-SNE projection shows visible distortion, with gaps that define clusters of points and give the impression that the original data has an inherently complex structure.

This visually salient distortion is barely captured by other quality metrics, as shown in \cref{fig:examples-gaps,fig:results-plane} (more examples can be found in the supplementary material), but it is significantly penalized by the GI.
Using the GI, a DR practitioner who encounters the two projections is able to confidently say that the t-SNE projection has important distortion; moreover, by visualizing the local distortions on top of the scatterplot, it becomes evident that the empty regions and dense clusters are artifacts of the DR method used to create the projection.

\begin{figure}[thbp]
    \centering
    
    \begin{subfigure}{0.32\columnwidth}
        \centering
        \includegraphics[height=2.5cm,keepaspectratio,alt={Projection of the Sphere dataset with mMDS with no distortion. The layout is uniform.}]{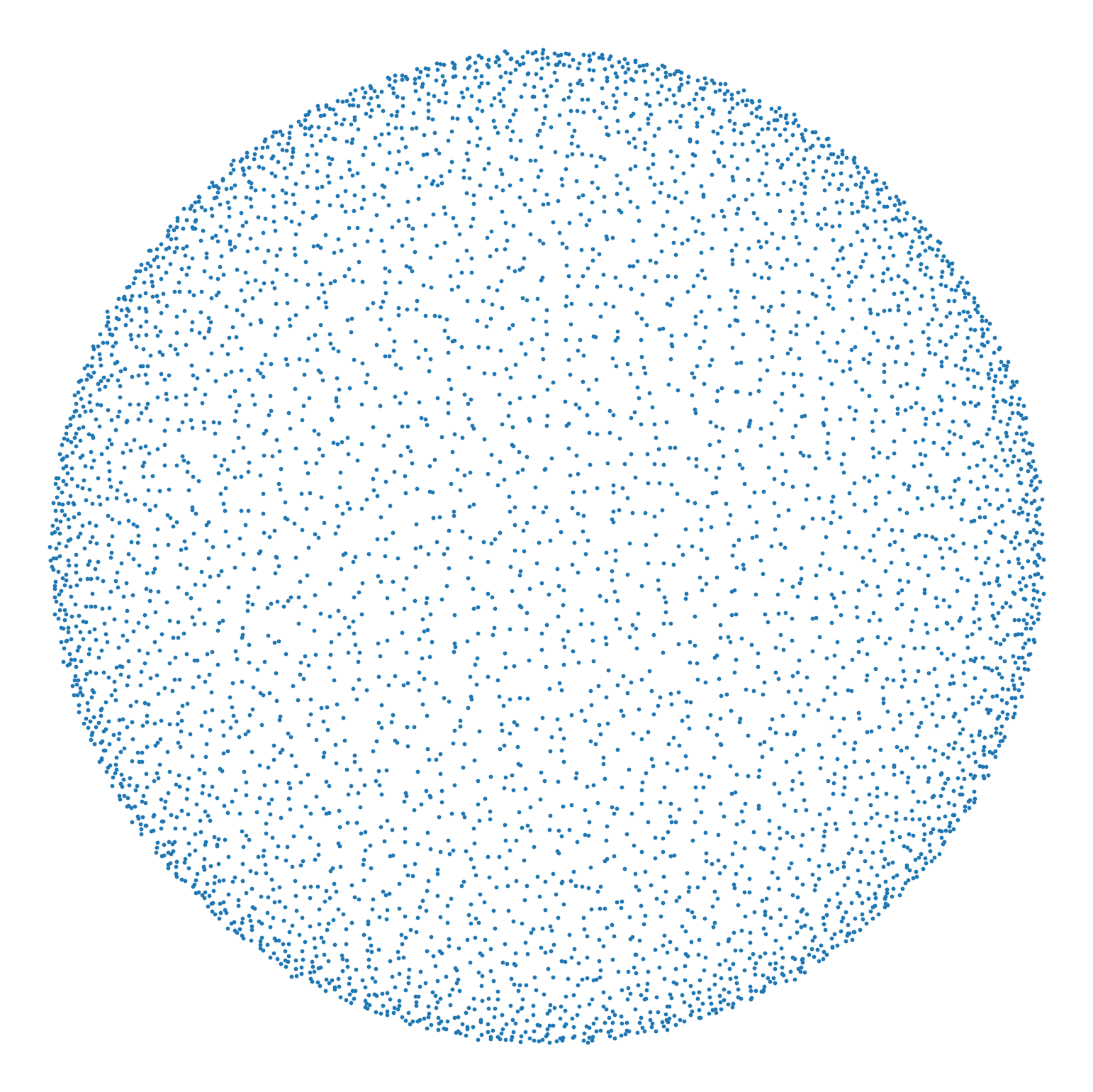}\\
        $\Delta = 0$\\
        \includegraphics[height=2.5cm,keepaspectratio,alt={Visualization of the GI on the previous layout. The inner part is slightly red, while the edge is slightly blue.}]{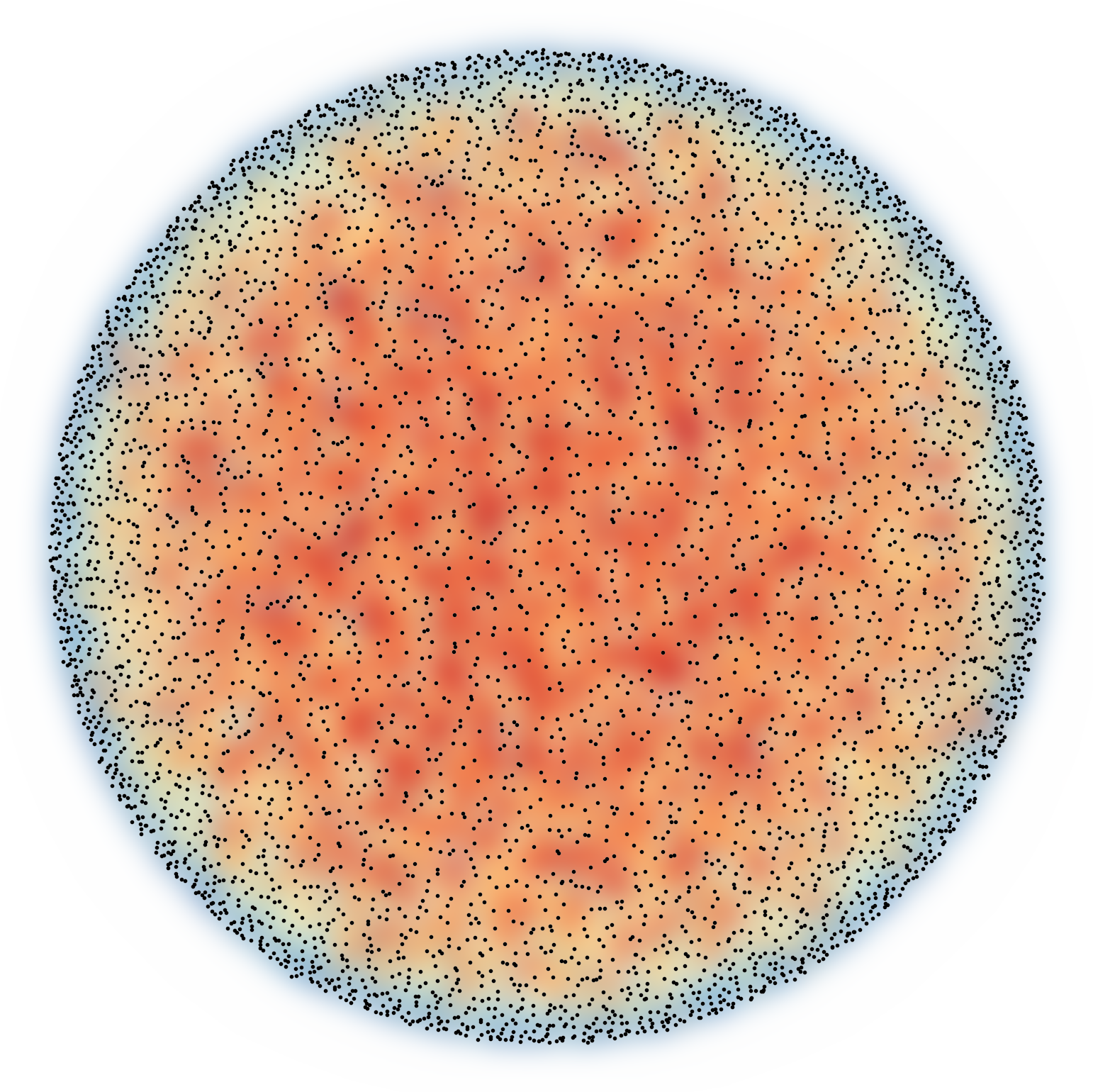}
    \end{subfigure}
    \begin{subfigure}{0.32\columnwidth}
        \centering
        \includegraphics[height=2.5cm,keepaspectratio,alt={Projection of the Sphere dataset with mMDS with no distortion. Gaps are created at random places.}]{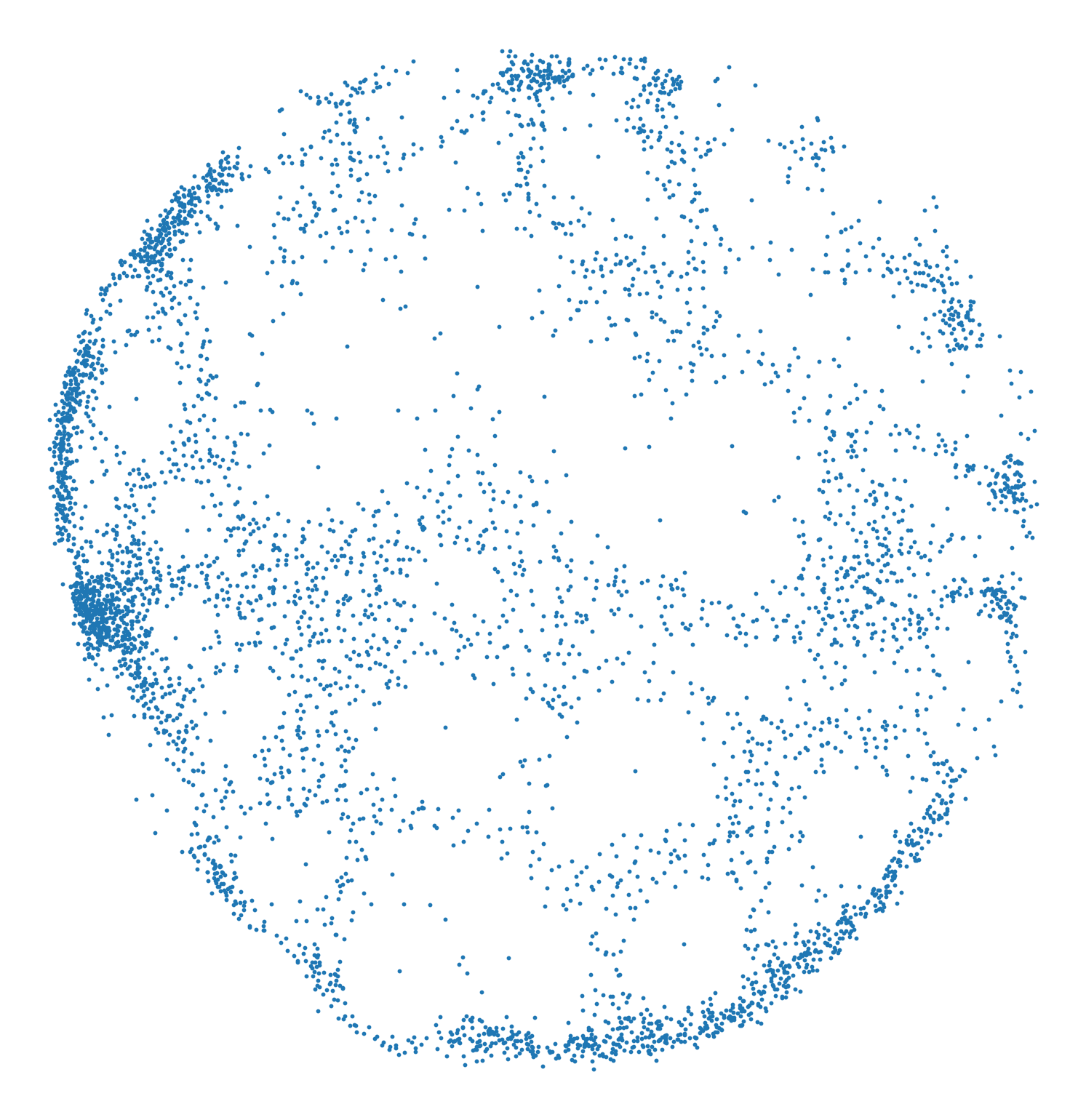}\\
        $\Delta = 0.1$\\
        \includegraphics[height=2.5cm,keepaspectratio,alt={Visualization of the GI on the previous layout. The holes are colored dark red, while clusters are blue.}]{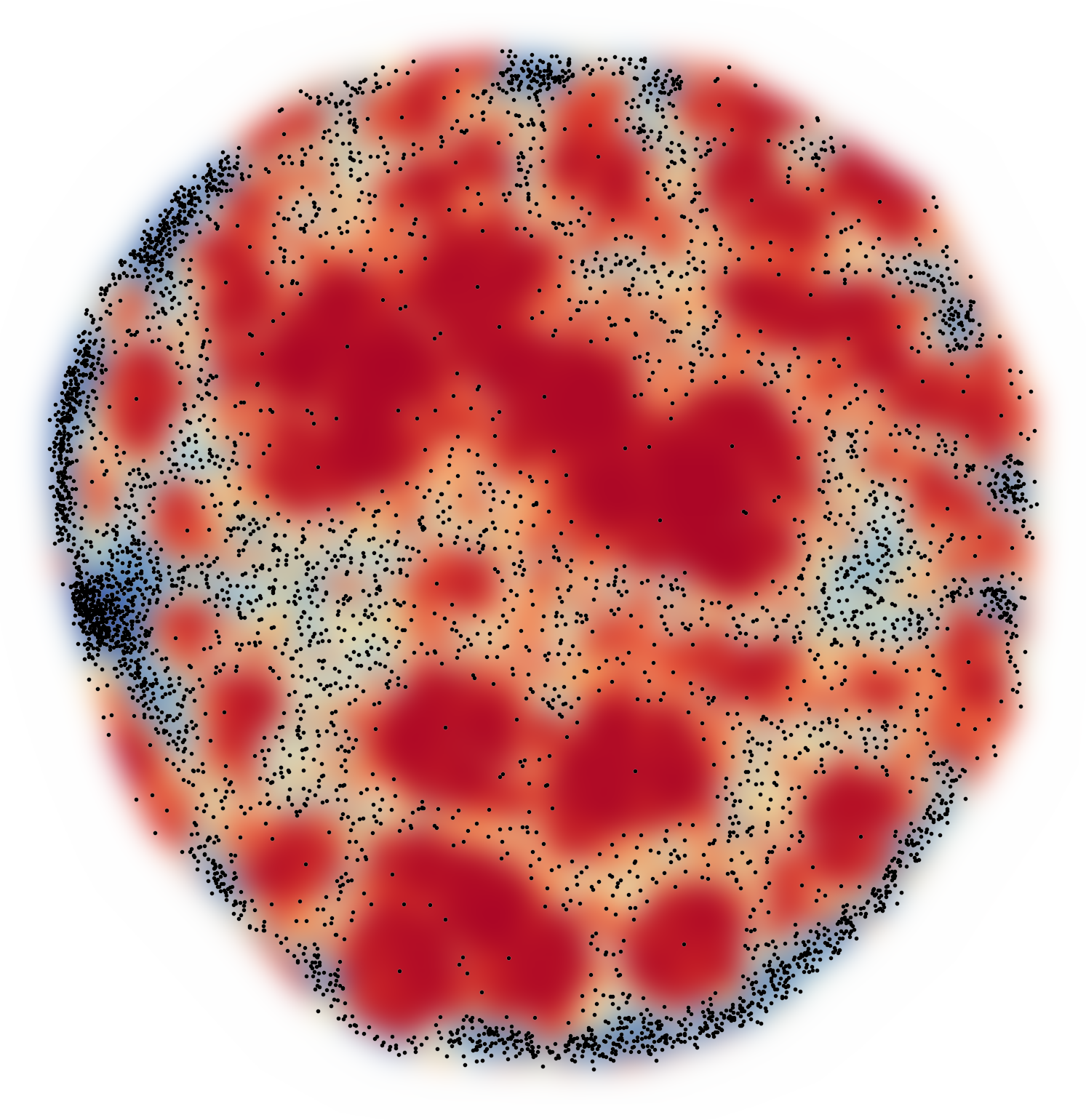}
    \end{subfigure}
    \begin{subfigure}{0.32\columnwidth}
        \centering
        \includegraphics[height=2.5cm,keepaspectratio,alt={Projection of the Sphere dataset with mMDS with no distortion. The previous gaps are increased in size.}]{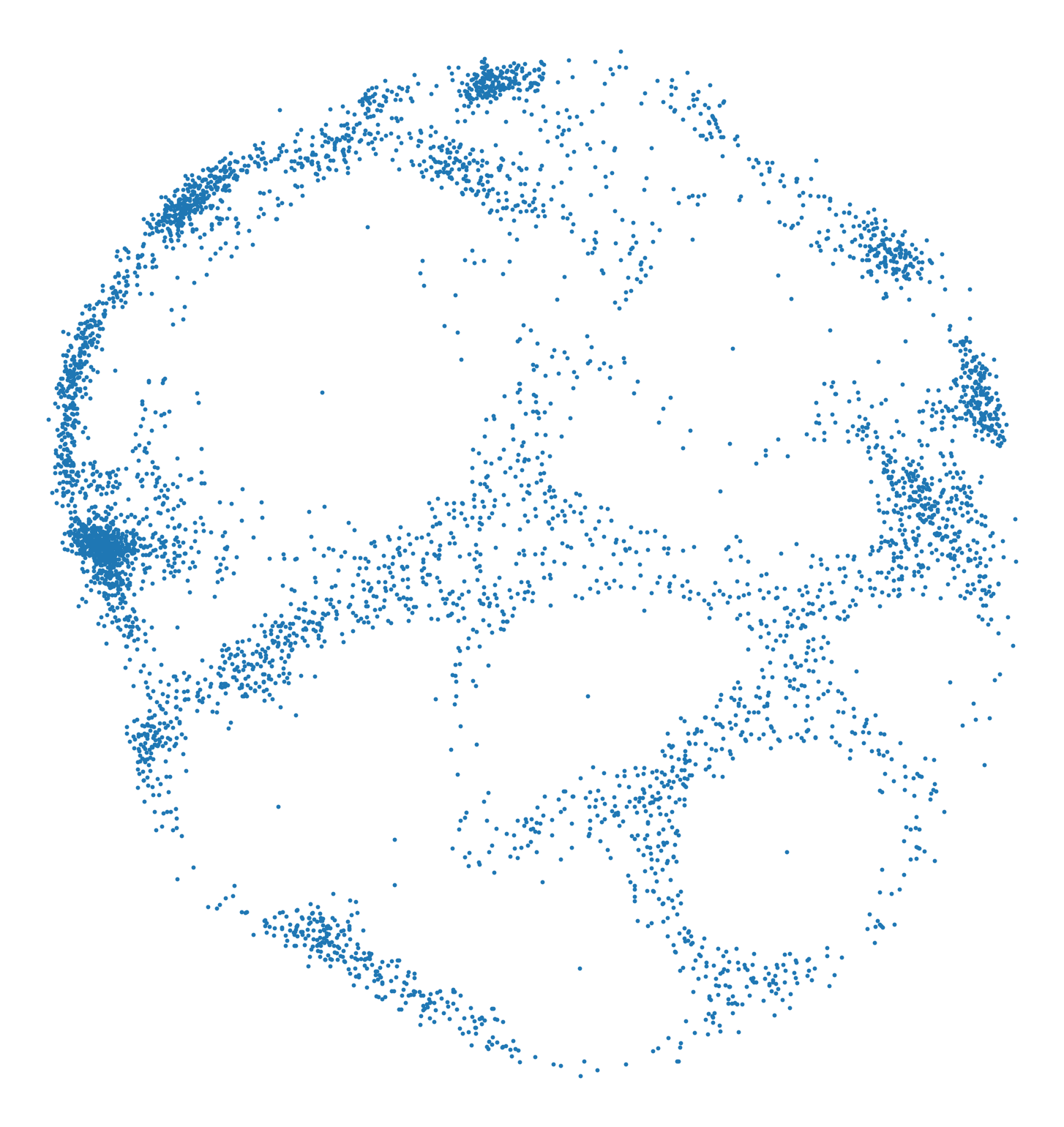}\\
        $\Delta = 0.2$\\
        \includegraphics[height=2.5cm,keepaspectratio,alt={Visualization of the GI on the previous layout. The holes are even darker red, while clusters are bluer.}]{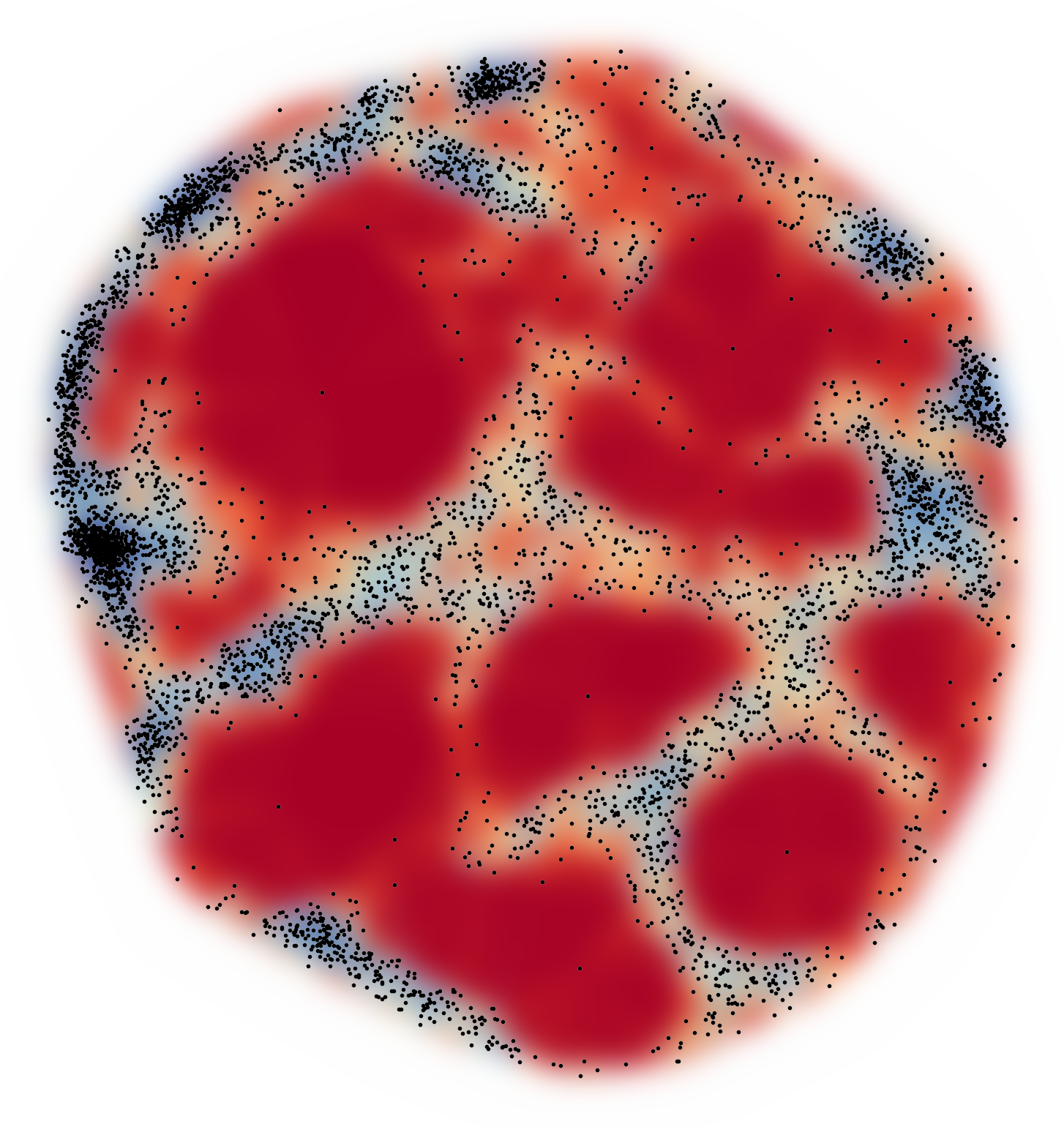}
    \end{subfigure}
    
    \vspace{0.5em}
    
    \begin{subfigure}{\columnwidth}
        \includegraphics[width=\linewidth,alt={Linechart of different metrics on the for increasing values of deformation. All the metrics remain mostly constant, except the GI, which increases smoothly from 0.444 to 0.745.}]{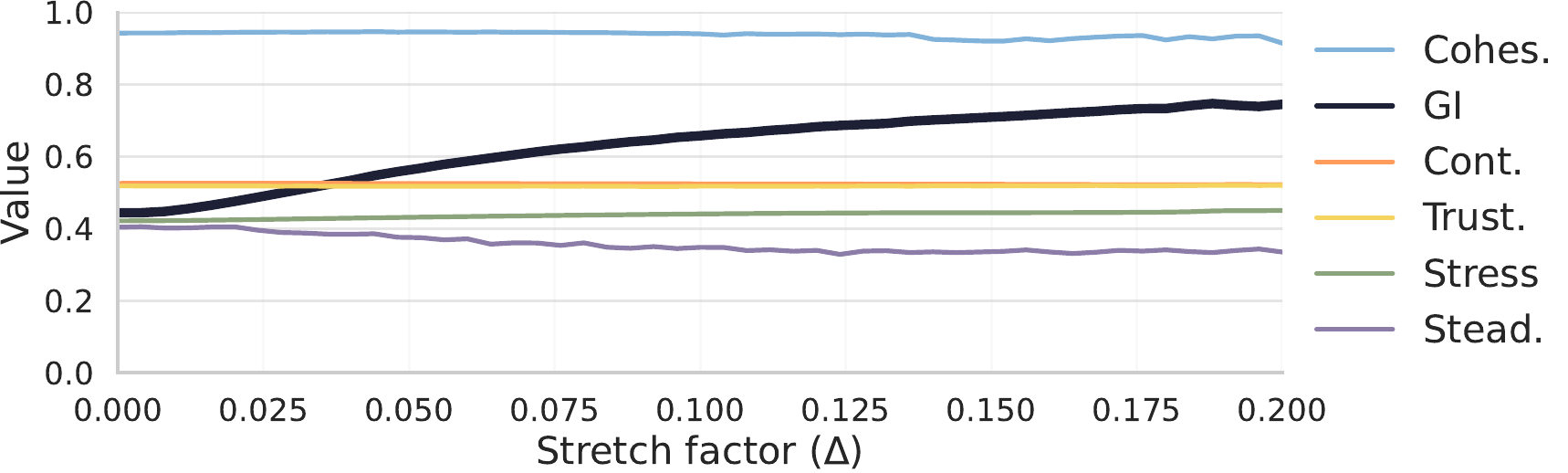}
    \end{subfigure}
    
    \caption{Behavior of multiple quality metrics at different levels of distortion of a mMDS projection of the \textit{sphere} dataset. Top: scatterplots of the dataset at different values of $\Delta \in [0,0.2]$; middle: visualization of the distortion with the GI for each of the previous scatterplots; bottom: line chart showing the value of each quality metric as $\Delta$ increases. Notice the overlap between continuity and trustworthiness. Note that all five reported metrics are in $[0,1]$: for stress and the GI, lower is better; for the rest, higher is better.}
    \label{fig:results-islands}
\end{figure}

For a more complex use case, we study the \textit{sphere} dataset: a set of $5$k points randomly placed inside a $200$-dimensional hypersphere of radius $1$.
The high intrinsic dimensionality of this dataset makes it difficult to project accurately to 2D, and some DR projections tend to create spurious clusters and gaps that, although visually salient, popular quality metrics cannot properly capture.
We consider the projection by metric Multidimensional Scaling~\cite{kruskal1964} (mMDS), shown in \cref{fig:results-islands} for $\Delta=0$. It captures the circular shape of the data, with a higher density towards the edges due to the distance concentration effect in high-dimensional space~\cite{peng2025}, and it does not show any spurious structure.
We distort this initial projection by iteratively optimizing a random subset of $50$ anchor points $a_1 \dots a_{50}$ and optimizing the rest of the 2D layout around them. This is done by displacing any other point $b \ne a_i$ in the direction of $\vec{a_ib}$ by a certain amount $\Delta$ ($\Delta=0$ indicates no displacement; $\Delta=1$ moves point $b$ until its distance in $\mathbb{X}$ matches that in $\mathbf{\hat{X}}$).
This type of distortion was studied before~\cite{ros2026_enhanced} and found to produce strong visual distortion not captured by popular quality metrics, even for high values of $\Delta$.

In the line chart in \cref{fig:results-islands}, the values of multiple quality metrics are shown as $\Delta$ (and thus distortion) increases. We see that for $\Delta=0$, most metrics agree on the existing level of distortion, since the data has already been projected from $200$ to $2$ dimensions. However, the surprising result occurs as the value of $\Delta$ increases, creating noticeable gaps in the layout, while most metrics remain constant.
Only the GI is capable of capturing this effect, increasing from $0.444$ for $\Delta=0$, to $0.745$ for $\Delta=0.2$. Additionally, the distortion plots of \cref{fig:results-islands} show that indeed the openings have been captured as a stretch of the empty space (red), while the clusters correspond to compressed areas (blue).

\begin{figure}[thbp]
    \centering
    
    \begin{subfigure}{0.32\columnwidth}
        \centering
        \includegraphics[height=2.5cm,keepaspectratio,alt={Projection of the Cube dataset with t-SNE (perplexity 20). The layout has wide gaps.}]{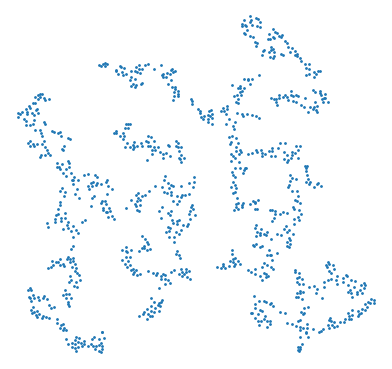}\\
        $p = 20$\\
        \includegraphics[height=2.5cm,keepaspectratio,alt={Visualization of the GI on the previous layout. The holes are colored dark red, while clusters are blue.}]{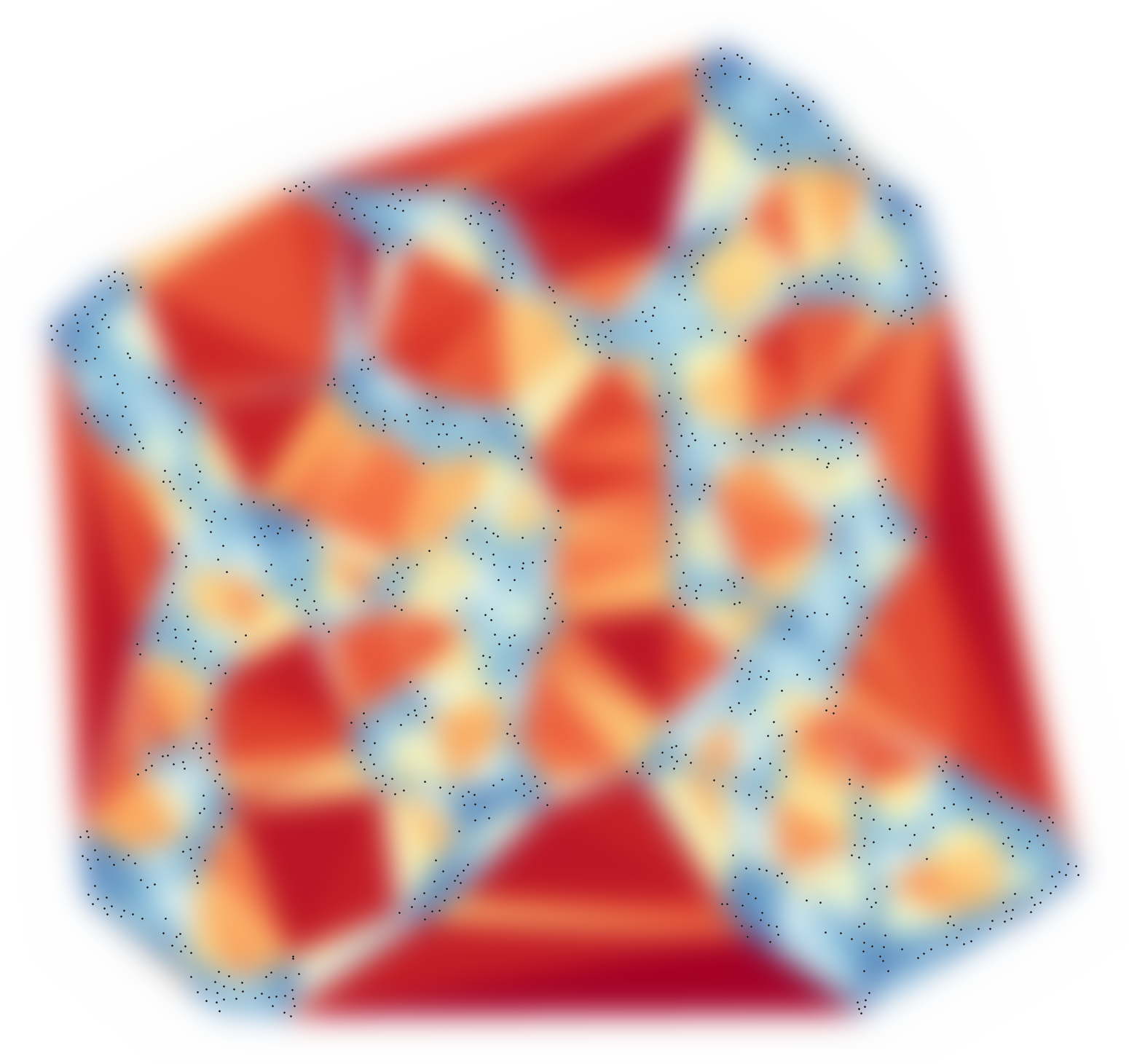}
    \end{subfigure}
    \begin{subfigure}{0.32\columnwidth}
        \centering
        \includegraphics[height=2.5cm,keepaspectratio,alt={Projection of the Cube dataset with t-SNE (perplexity 70). The layout resembles the previous one, but points are distributed more uniformly.}]{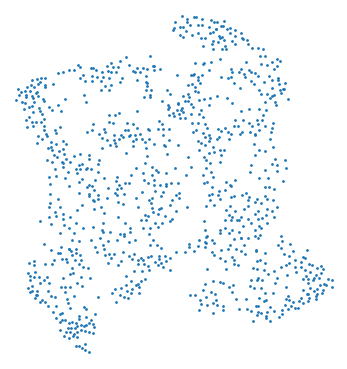}\\
        $p = 70$\\
        \includegraphics[height=2.5cm,keepaspectratio,alt={Visualization of the GI on the previous layout. The smaller gaps are less red.}]{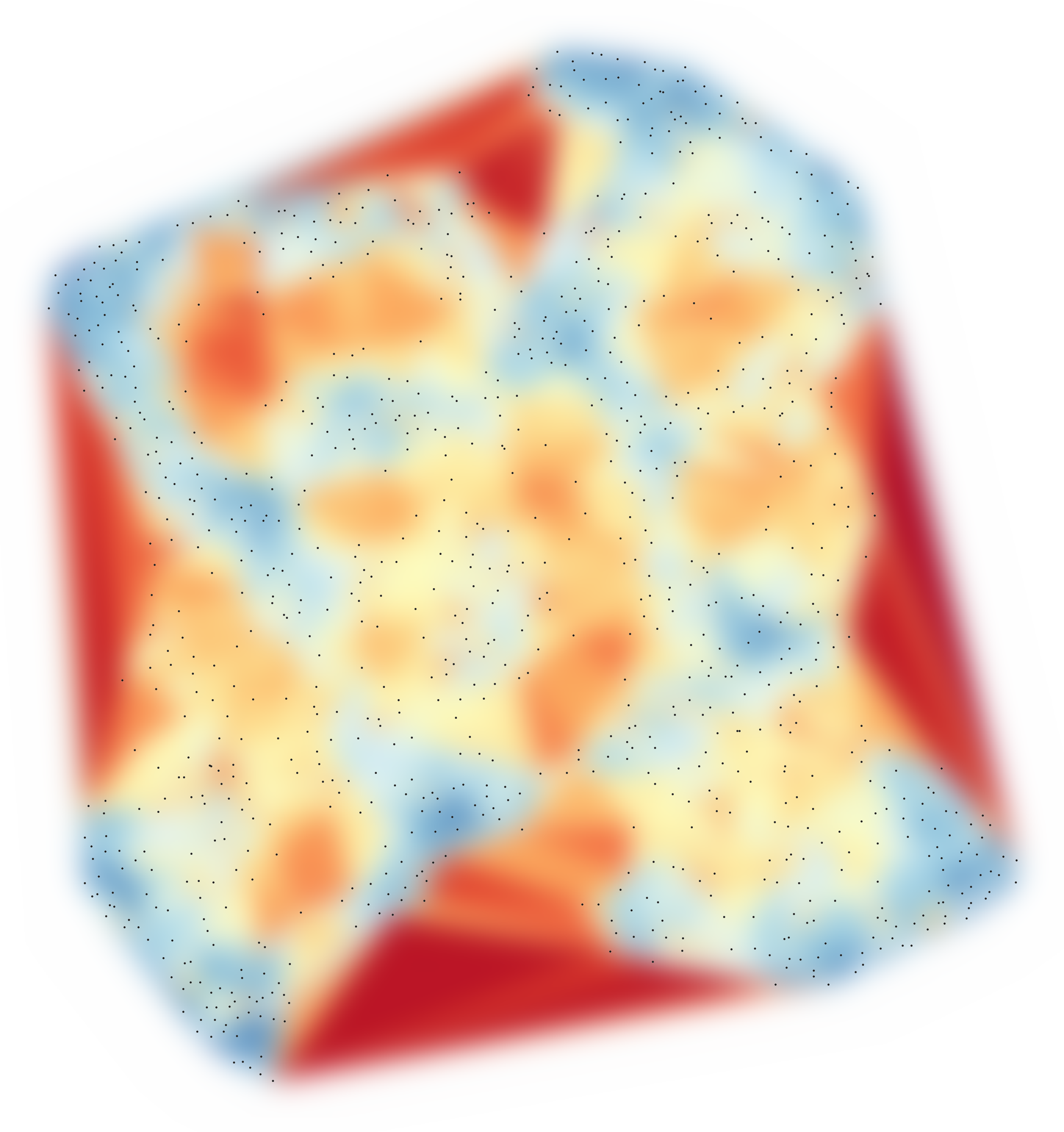}
    \end{subfigure}
    \begin{subfigure}{0.32\columnwidth}
        \centering
        \includegraphics[height=2.5cm,keepaspectratio,alt={Projection of the Cube dataset with t-SNE (perplexity 300). The layout is mostly uniform.}]{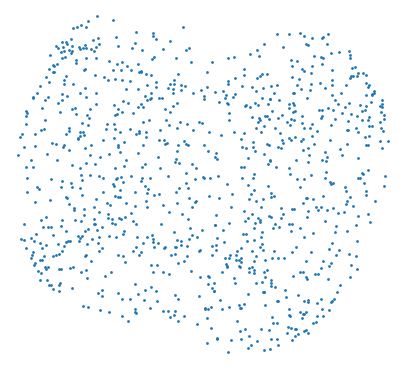}\\
        $p = 300$\\
        \includegraphics[height=2.5cm,keepaspectratio,alt={Visualization of the GI on the previous layout. The shape of the flattened cube is visible.}]{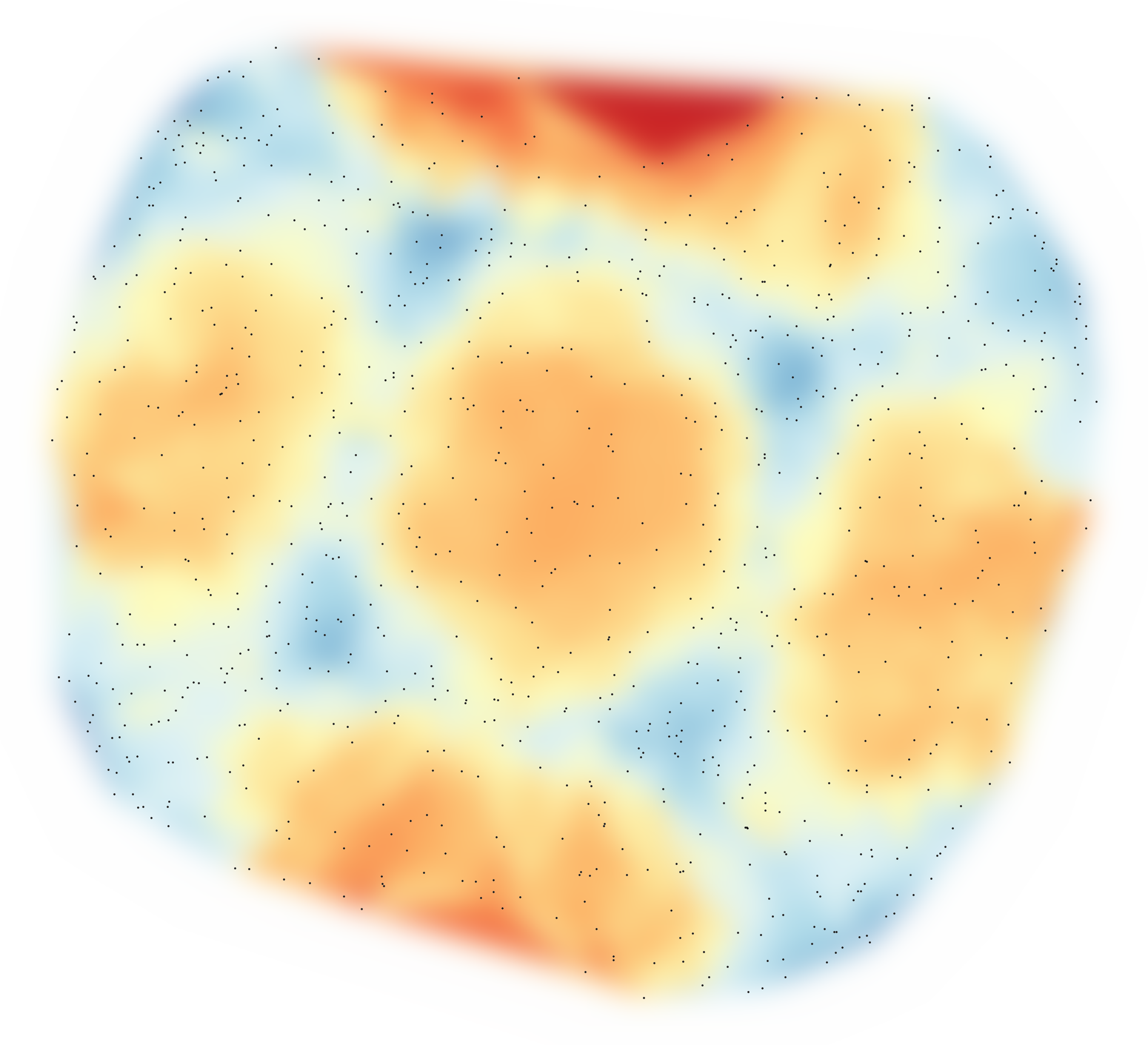}
    \end{subfigure}
    
    \vspace{0.5em}
    
    \begin{subfigure}{\columnwidth}
        \includegraphics[width=\linewidth,alt={Linechart of different metrics on the for increasing values of deformation. Most the metrics remain constant, except the GI (decreases) and cohesiveness (increases).}]{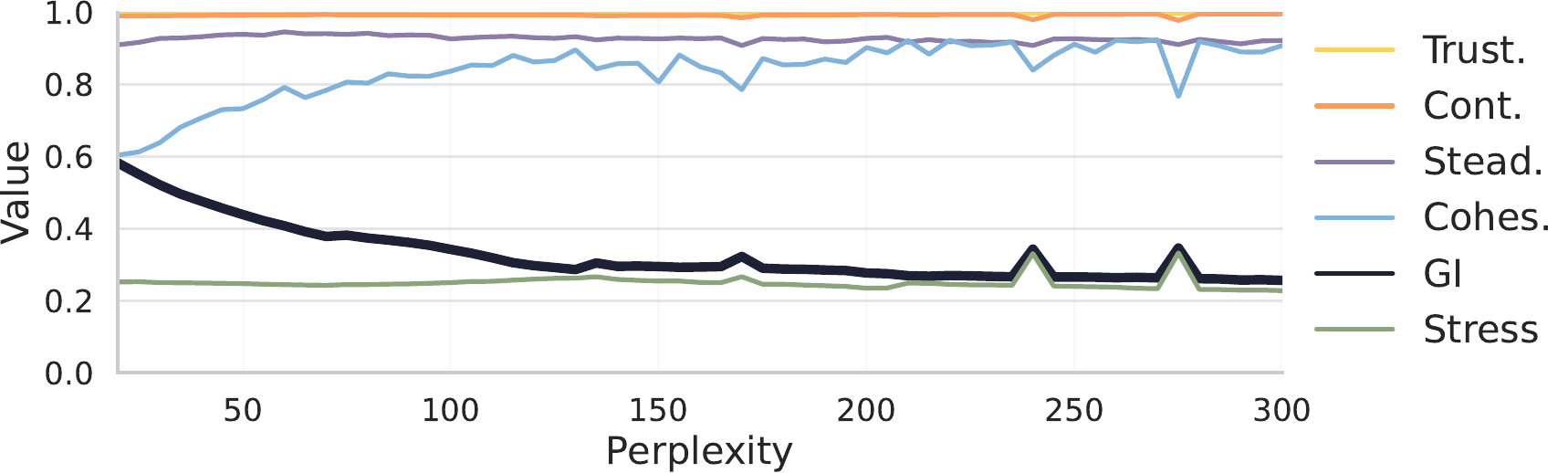}
    \end{subfigure}
    
    \caption{Behavior of multiple quality metrics for t-SNE projections of the \textit{cube} dataset, with increasing values of perplexity $p \in [20,300]$. Top: scatterplots of the dataset at different values of $p$; middle: visualization of the distortion with the GI for each of the previous scatterplots; bottom: line chart showing the value of each quality metric as $p$ increases. Notice the overlap between continuity and trustworthiness. Note that all five reported metrics are in $[0,1]$: for stress and the GI, lower is better; for the rest, higher is better.}
    \label{fig:results-perplexity}
\end{figure}

Finally, \cref{fig:results-perplexity} shows the GI behavior on t-SNE projections of the \textit{cube} dataset (already introduced in \cref{sec:interpreting}, with its PCA projection shown in \cref{fig:results-interpreting-cube}).
As the perplexity $p$ increases (between $p=20$ and $p=300$, at intervals of $5$), the global structure of the data is better captured:
with $p=20$, spurious patterns and gaps emerge, which are heavily penalized by the GI; for $p=70$ the cube appears ``unfolded'' in the 2D plane, with the GI mostly penalizing the gaps on the sides, corresponding to the cuts of the original 3D shape; finally, for $p=300$ the scatterplot presents a mostly uniform and structureless set of points, but visualizing the GI reveals how it maps to the original 3D shape, with some regions of points having been stretched, and other compressed.

The comparison with other metrics in the line chart of \cref{fig:results-perplexity} shows that only cohesiveness significantly captures the difference between projections, although its values are noisier.
The outliers for $p=240$ and $p=275$ are due to a misplaced cluster in the layout (creating a distorted visual gap); these examples, along with additional studies from other datasets, can be found in the supplementary material.

\subsection{GI to visualize local distortion}

\begin{figure}[thbp]
    \centering
    \begin{subfigure}{0.32\columnwidth}
        \includegraphics[width=\linewidth,alt={Visualization of per-point-stress coloring points. There are regions with more and less stress, but they have no relation to the layout.}]{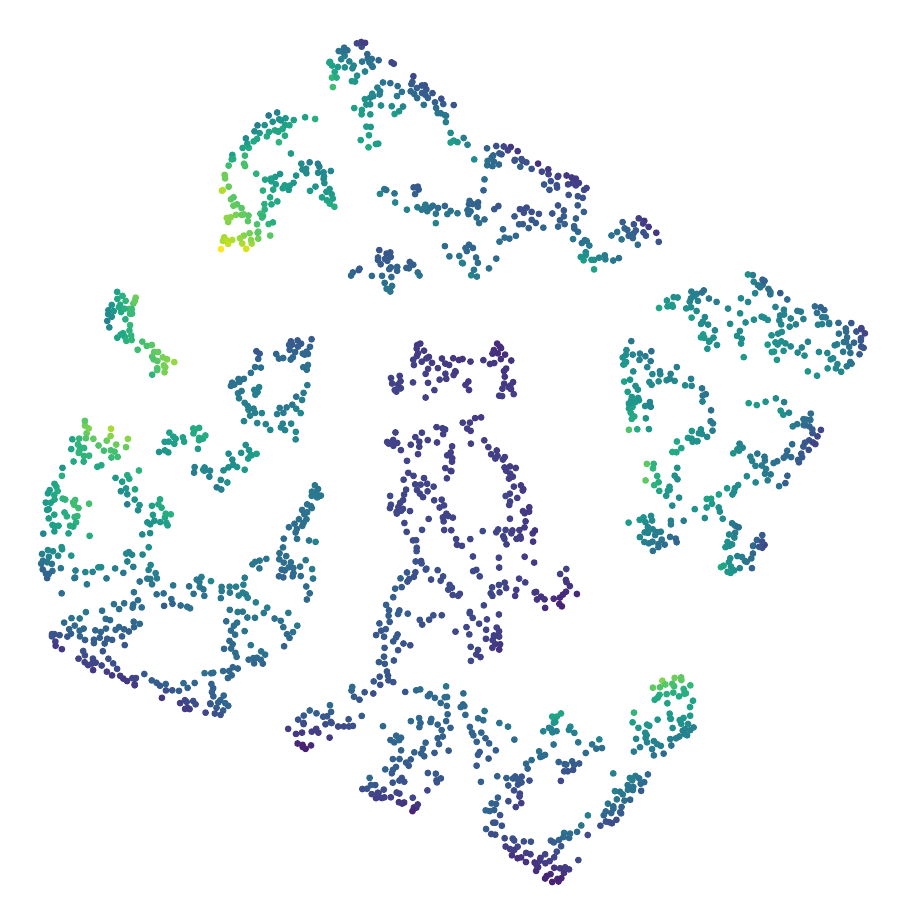}
        \caption{}
        \label{fig:results-visualization-points}
    \end{subfigure}
    \hfill
    \begin{subfigure}{0.32\columnwidth}
        \includegraphics[width=\linewidth,alt={Visualization of per-point-stress through Voronoi cells. The stress values are the same as the previous image.}]{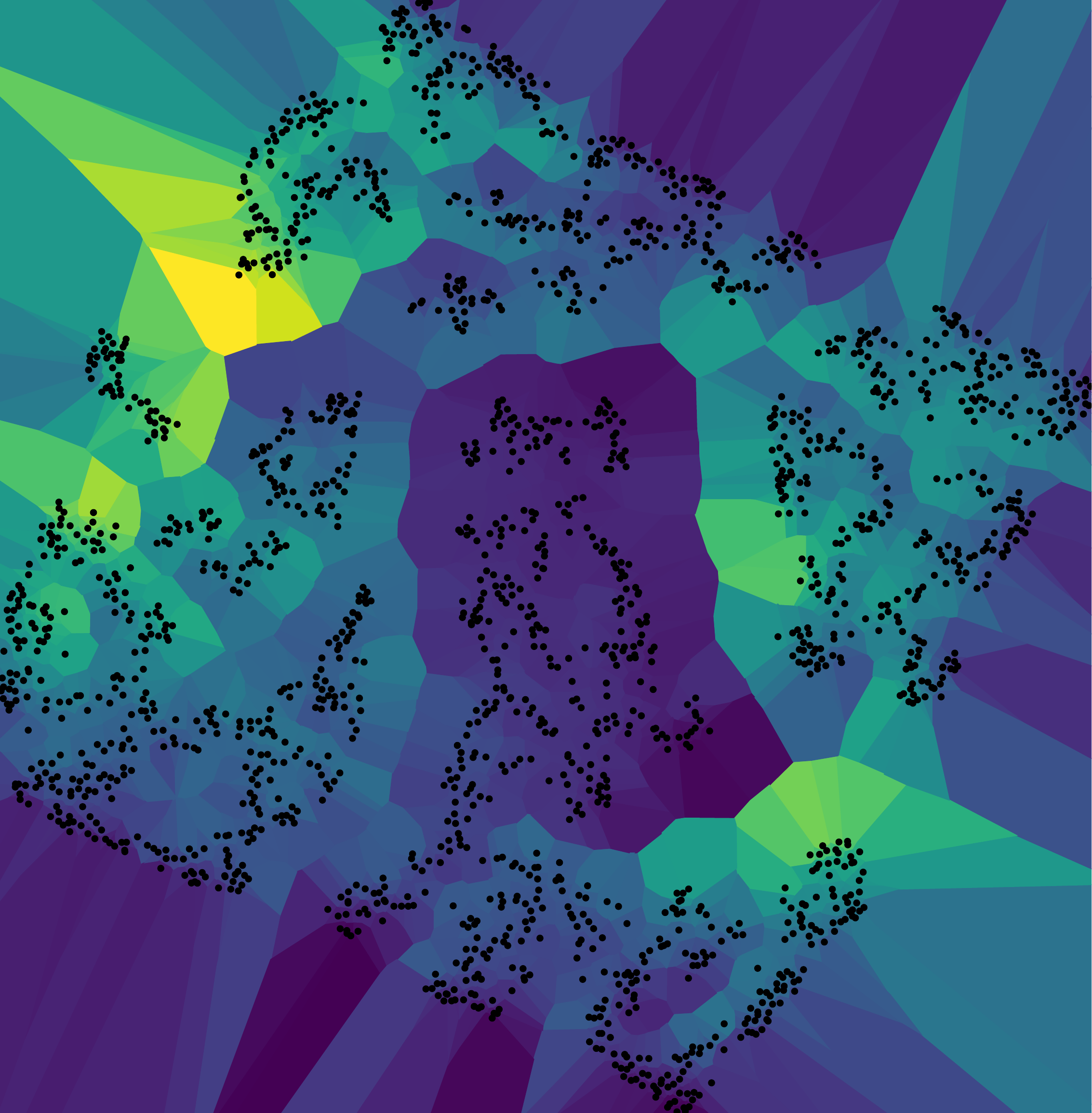}
        \caption{}
        \label{fig:results-visualization-voronoi}
    \end{subfigure}
    \hfill
    \begin{subfigure}{0.32\columnwidth}
        \includegraphics[width=\linewidth,alt={Visualization of per-point-stress through triangles. The values change smoothly, but are not informative about the distortion.}]{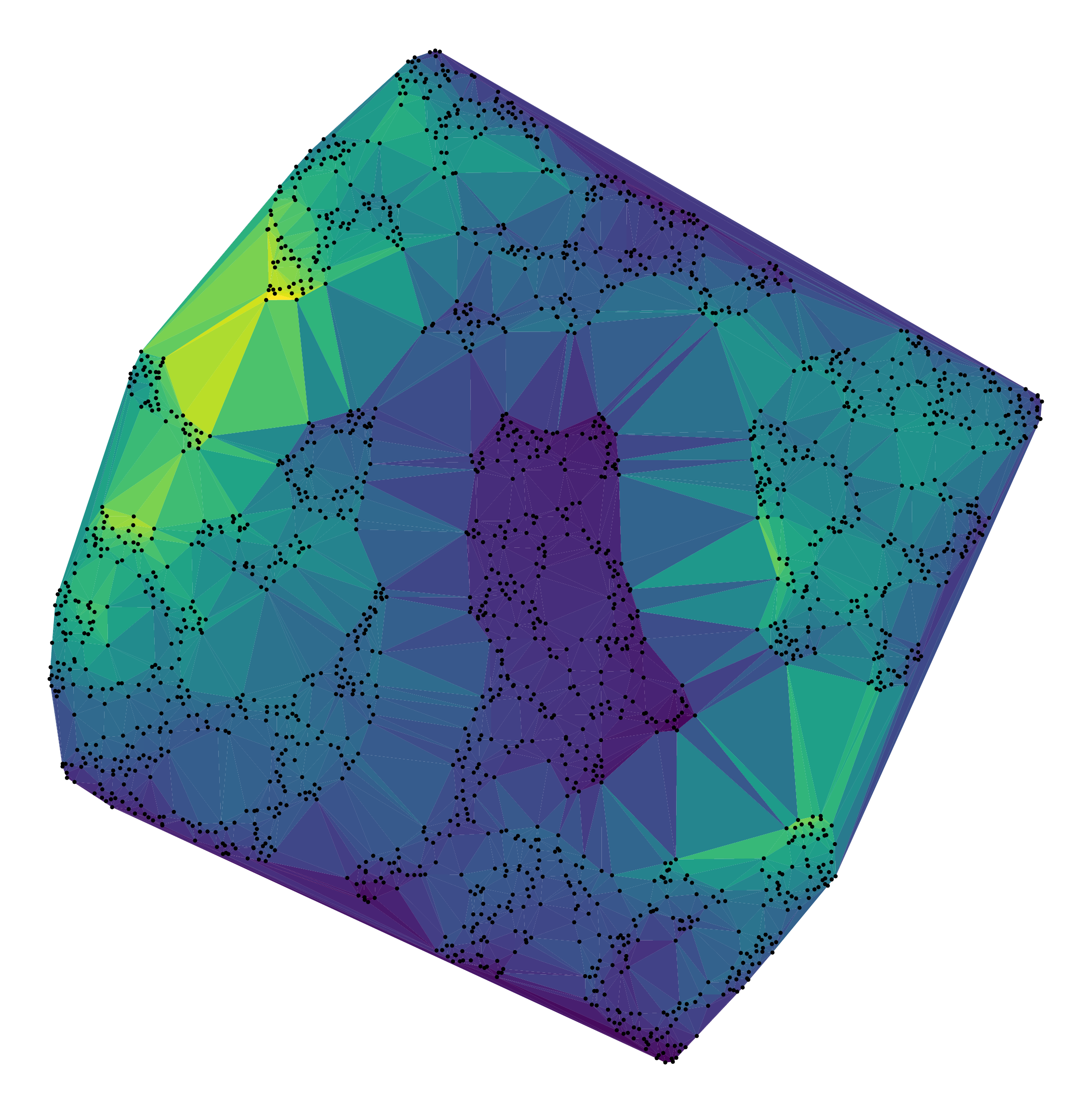}
        \caption{}
        \label{fig:results-visualization-triangles}
    \end{subfigure}
    \caption{Per-point stress of a t-SNE projection of the \textit{plane} dataset visualized with different coloring approaches: (a) points, (b) Voronoi cells, (c) Delaunay triangles. Colors go from purple (lowest) to yellow (highest).}
    \label{fig:results-visualization}
\end{figure}

Previous results compared quality metrics that capture the overall distortion of the projection as a single scalar value. However, another common scenario is to use them to visualize local distortion, as discussed in \cref{sec:rw-visualization}.

As mentioned throughout the paper, most quality metrics are computed per-point, which makes local distortion in the empty space hard to interpret.
In \cref{fig:results-visualization}, we show three different strategies for visualizing local distortion. In this example, we use per-point stress, but note that this applies to any per-point scalar metric.
In~\subref{fig:results-visualization-points}, points are colored according to the value of the metric; while this approach is common due to its simplicity and allows for visualizing regional patterns of distortion, users are not provided with any information about the distortion in the empty regions.
In~\subref{fig:results-visualization-voronoi}, the corresponding Voronoi cell of each point is colored according to its stress value; this approach, popularized by CheckViz~\cite{Lespinats_Aupetit_2011}, displays per-point distortion in the background, but suffers from differences in cell size (some points are more salient) and hard boundaries between cells in the middle of empty regions.
In~\subref{fig:results-visualization-triangles}, Delaunay triangles are colored based on the average values of their three vertices; while patterns are smoother than in the Voronoi diagram, averaging the three vertex values can obscure the interpretation of the metric, especially for large triangles with vertices in distant clusters.

Through these examples, the need for a metric that measures distortion in the empty space becomes evident. Not only is stress unable to correctly identify the local distortion patterns (compare with the GI plot on \cref{fig:results-plane}b), but any attempt to color the background according to a per-point metric can potentially mislead the user into interpreting it as a distortion of space.

\subsection{Using the GI on large datasets} \label{sec:results-scalability}

When using DR in a practical setting, one can encounter large datasets with thousands to millions of points.
In such cases, fast quality metrics are needed as they allow evaluation of the distortion of the projection in a reasonable time and without excessive computational resources.

\begin{table}[thb]
\caption{Execution time (in seconds) for different quality metrics on PCA projections of datasets with an increasing size (defined by the number of points $N$).
The experiments have been run on a mid-range laptop with an \textit{Intel Core i7-13700H} CPU and 16 GiB of memory.
The reported times are the average over $10$ iterations. Missing values indicate that the metric could not be computed due to excessive memory requirements.}
\label{tab:time}
\centering
\begin{tabular}{ccccc}
\textbf{Metric}                               & \begin{tabular}[c]{@{}c@{}}\textit{Plane}\\ $N=2$k\end{tabular} & \begin{tabular}[c]{@{}c@{}}\textit{Sphere}\\ $N=5$k\end{tabular} & \begin{tabular}[c]{@{}c@{}}\textit{Fiber (small)}\\ $N=19$k\end{tabular} & \begin{tabular}[c]{@{}c@{}}\textit{Fiber (large)}\\ $N=250$k\end{tabular} \\ \hline
\multicolumn{1}{c|}{\textbf{Stress}}          & 0.03                                                 & 0.88                                                  & 4.39                                                    & -                                                      \\
\multicolumn{1}{c|}{\textbf{Trust.}} & 0.19                                                 & 1.95                                                  & 21.9                                                    & -                                                      \\
\multicolumn{1}{c|}{\textbf{S\&C}}            & 0.91                                                 & 45.69                                                 & -                                                       & -                                                      \\
\multicolumn{1}{c|}{\textbf{GI}}              & 0.07                                                 & 0.19                                                  & 0.72                                                    & 9.75                                                  
\end{tabular}
\end{table}

\Cref{tab:time} shows a comparison of execution time for different metrics as the size of the data increases.
The \textit{plane} and \textit{sphere} datasets, which we consider medium-sized, were introduced in previous sections.
To test the scalability of quality metrics on a large dataset, we used the \textit{fiber} dataset~\cite{Poco_2012}, which consists of $250$k brain-tracking fiber streamlines; the \textit{small} version is a subset of labeled data from the \textit{big} one.
We limit our study to a dataset of this magnitude, as we are already unable to compute any of the other metrics to compare to the GI.

Global metrics such as stress have complexity $O(N^2)$ due to the need to compute all pairwise distances.
Trustworthiness, a local metric, relies on nearest-neighbor queries; since acceleration structures such as kd-trees~\cite{maneewongvatana1999} are inefficient in high-dimensional spaces, a brute-force approach is used.
Thus, these two types of metrics exhibit similar scalability and can be used on datasets with up to a few tens of thousands of points.
Unsupervised cluster-level metrics rely on expensive clustering algorithms and exhibit the worst scalability among all the metrics tested. Steadiness and cohesiveness (S\&C) use the same clustering result from HDBSCAN~\cite{mcinnes2017} and are therefore computed together.

As mentioned in \cref{sec:met-triangulation}, the GI has a cost of $O(N \log N)$ ($N$ being the number of points in $\mathbf{X}$), determined by the Delaunay triangulation.
It shows the best scalability of all metrics tested, both in terms of execution time and memory usage; it achieved an average time under a second for the \textit{fiber (small)} dataset and it was the only metric capable of processing the largest dataset with our available computational resources, in around $10$ seconds and with a memory usage of less than $2$ GiB.
Thus, it is a viable option to evaluate the quality of projections with a large number of points, even when other metrics fail.

\subsection{Evaluating the Delaunay triangulation}

In \cref{sec:met-triangulation}, we motivated the use of the Delaunay triangulation.
However, a consequence of this choice is that the final value of the GI is not continuous with respect to the placement of the points in $\mathbf{X}$, since a small change can create a different triangulation, which in turn would produce different values of triangle deformation and change the final aggregated result.
Note that this effect is also present in most local and cluster-level metrics that depend on discrete structures such as neighborhoods (e.g., trustworthiness and continuity) and clusters (e.g., steadiness and cohesiveness). 
Although the GI has shown stability in the controlled cases of \cref{sec:results-overall}, in this section we study its effect on more realistic examples.

\begin{figure}
    \begin{subfigure}{\columnwidth}
        \includegraphics[width=.3\linewidth,alt={t-SNE projection of the COIL20 dataset. Many patterns are visible.}]{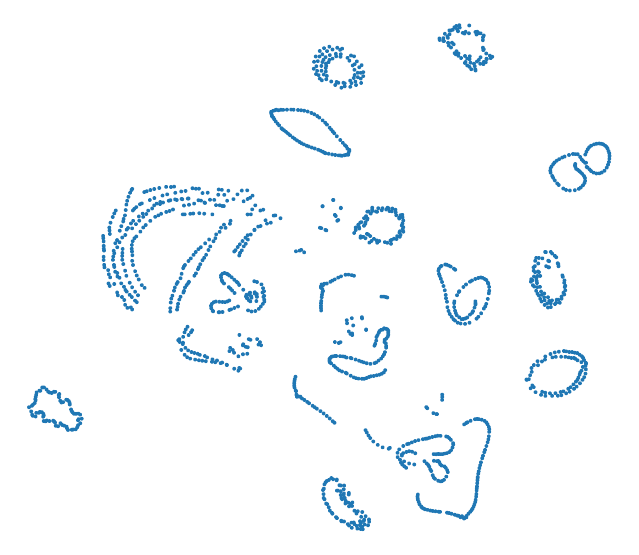}
        \hfill
        \includegraphics[width=.3\linewidth,alt={t-SNE projection of the COIL20 dataset with some random noise. Cluuters are still distinguishable.}]{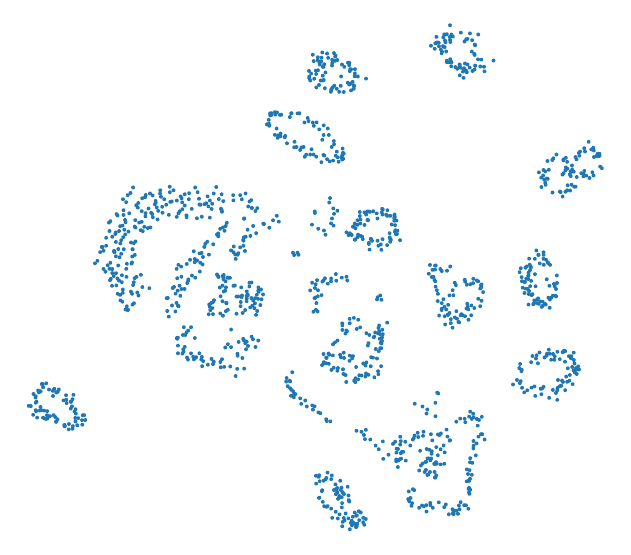}
        \hfill
        \includegraphics[width=.3\linewidth,alt={t-SNE projection of the COIL20 dataset with more random noise. The original structure is barely recognizable.}]{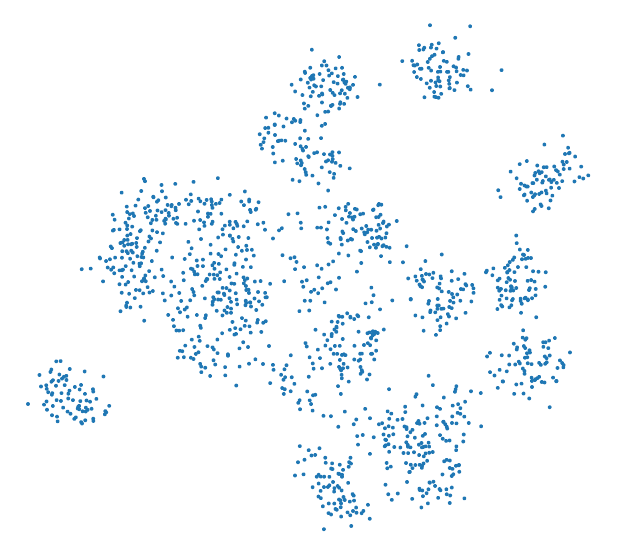}
    \end{subfigure}
    
    \noindent
    \makebox[\columnwidth]{%
        \makebox[0.3\linewidth]{\centering \small $\eta = 0$}%
        \hfill
        \makebox[0.3\linewidth]{\centering \small $\eta = 0.005$}%
        \hfill
        \makebox[0.3\linewidth]{\centering \small $\eta = 0.02$}%
    }

    \vspace{1em}
    
    \begin{subfigure}{\columnwidth}
        \centering
        \includegraphics[width=.48\linewidth,alt={Boxplot of standard deviation for the GI. It increases proportionally to the noise.}]{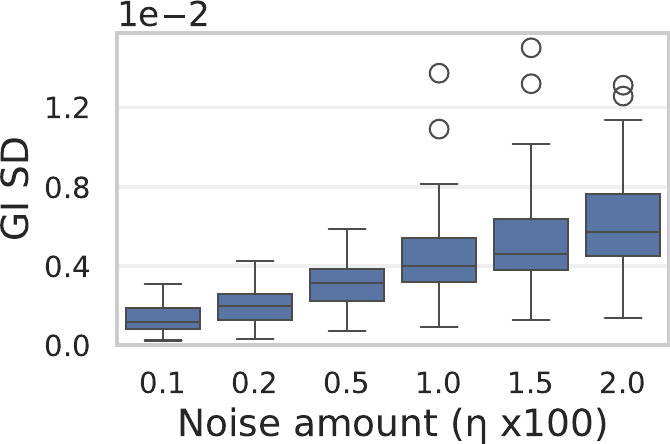}
        \hfill
        \includegraphics[width=.48\linewidth,alt={Boxplot of standard deviation for stress. It increases proportionally to the noise.}]{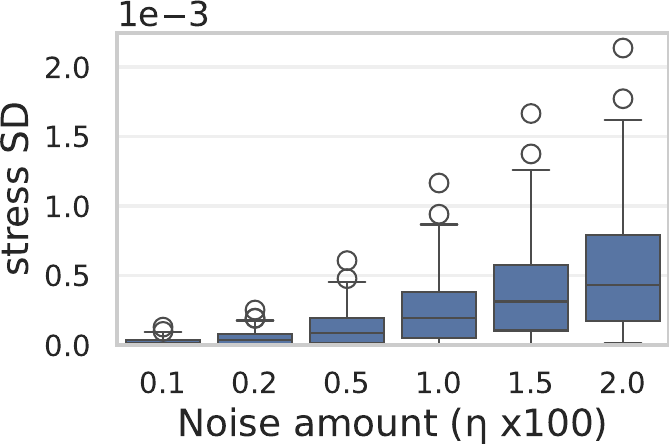}
    \end{subfigure}
    \caption{Boxplots of the standard deviation (SD) of the GI (left) and stress (right). Each bar corresponds to a level of noise $\eta \in [0.001,0.02]$, computed over $18 \times 3$ different datasets. The SD has been computed over the values of each quality metric on $10$ repetitions of the same layout with random noise. At the top, we include, as an example, t-SNE projections of the \textit{COIL20} dataset with different levels of noise.}
    \label{fig:boxplots-noise}
\end{figure}

To this end, we have used the diverse set of 18 datasets from Espadoto et al.~\cite{Espadoto_Martins_Kerren_Hirata_Telea_2021}, which feature different data types, sizes (both the number of points and the number of dimensions), and other particularities.
For each, we computed projections using PCA, t-SNE, and mMDS.
Note that the \textit{orl} dataset contained errors in the original publication, but these have been fixed prior to the computation of the results presented here.
On each of the $18 \times 3$ projections, we measure the standard deviation of the GI value as we randomly jitter the points over $10$ iterations.
For comparison, we have run the same analysis with stress, which is differentiable with respect to the point positions, and thus we expect it to behave consistently in this test.
Note that the 2D layouts have been uniformly scaled to $[0,1]^2$ to ensure that noise of the same magnitude has a similar effect across all of them; since both the GI and stress are scale-invariant, this change has no effect on the final values of the metrics.
We also include the projections of the $18$ datasets with different DR methods, along with reported quality values of multiple metrics, in the supplementary material.

The results are shown in the form of a boxplot in \cref{fig:boxplots-noise}.
As expected, the standard deviation increases with the noise level; however, it does so consistently and within reasonable values.
The higher standard deviation values for the GI can also be attributed to its higher sensitivity to small visual changes compared to stress, as shown previously.

\subsection{A practical use case for the GI}

To complement the previous results, we present an example of the use of the GI in the context of explainable artificial intelligence (XAI).
We trained a small convolutional neural network (CNN) to identify digits from the MNIST dataset~\cite{lecun2010mnist} ($28 \times 28$ grayscale images of handwritten digits); the model achieves an accuracy of $99$\%, but it remains a black box to users. A common XAI practice to gain insight into the behavior of the network is to visualize its latent states; since these are usually of high-dimensionality, DR techniques are used~\cite{rauber2017}.
In our example, we consider a t-SNE projection of the last hidden layer of the CNN, which consists of $64$-dimensional embeddings of the test instances; this is shown in \cref{fig:usecase}A.
In the supplementary material, we provide more information about the dataset, the architecture of the CNN, and additional results with alternative models and datasets.

\begin{figure}
    \centering
    \includegraphics[width=\linewidth,alt={Visualization of the latent space of a CNN with a t-SNE projection, with 10 visible clusters, uniformly shaped and spaced. The visualization of the GI shows most clusters in blue, but some are not; space between clusters is mostly dark red, with less intensity between some pairs.}]{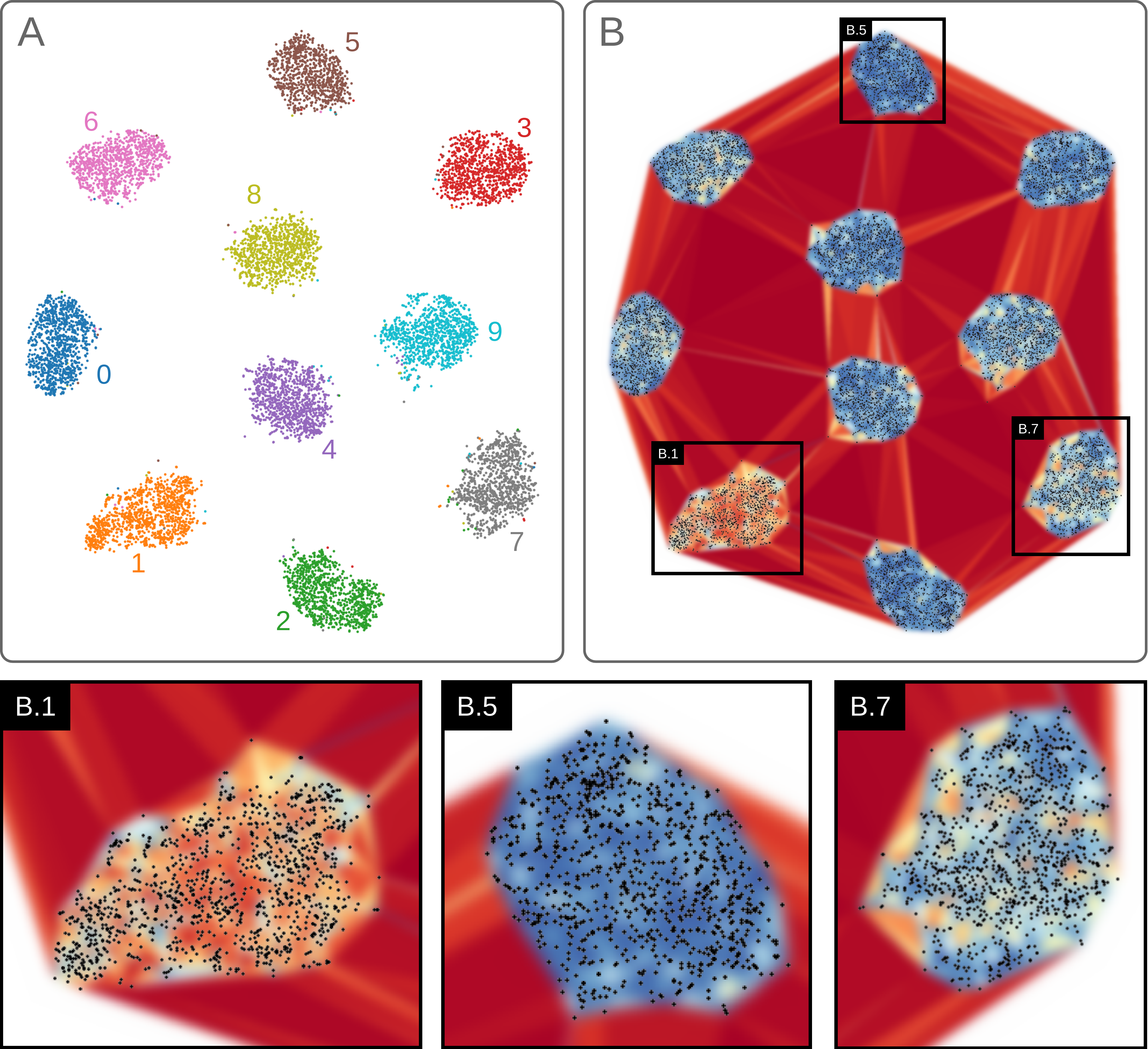}
    \caption{t-SNE projection of CNN embeddings of the MNIST dataset ($1$k instances for each class \textit{0}-\textit{9}; $10$k in total). (A) Scatterplot of the projection with points colored according to their label. (B) Corresponding GI visualization, with clusters \textit{1}, \textit{5}, and \textit{7} enlarged at the bottom.}
    \label{fig:usecase}
\end{figure}

\Cref{fig:usecase}A shows a scatterplot of the projection, with points colored according to their label.
It is expected that a neural network with $99$\% accuracy will have well-separated embeddings in its last hidden layer, and the DR plot confirms this, presenting $10$ distinct clusters, each corresponding to a single class (with some exceptions of misclassified instances).
All clusters present a similar compact shape, and their global position is generally meaningless, as expected from a local technique (t-SNE with a perplexity value of $100$).
However, a more complete picture emerges when visualizing the GI of the projection (\cref{fig:usecase}B).

The first observation is that clusters are compressed, while the empty space between them is heavily stretched. This is common and expected for GI visualizations of local DR methods, such as t-SNE. Although experienced DR practitioners are likely aware of this behavior, it can help avoid misuse of these techniques by less familiar users~\cite{jeon2025}.
This large distortion of the empty space is heavily penalized when aggregating the GI, which results in a value of $0.81$ (compared to a better $0.63$ for mMDS, a global technique), yet t-SNE's clearly separated clusters may still be preferred in many analytical tasks, and visualizing the GI can provide further insights to complement it.
On a global scale, it provides some information about the positioning of the clusters. More intense red indicates that the distance between two clusters has been more stretched than other pairs; thus, we can infer that cluster pairs (\textit{5},\textit{8}), (\textit{0},\textit{6}), (\textit{0},\textit{8}), or (\textit{4},\textit{9}) are in reality closer than they appear in the t-SNE projection. This is reasonable, given the similar patterns of their handwritten shapes.
In contrast, other pairs of clusters have been \emph{less stretched}; examples are pairs (\textit{4},\textit{8}) and (\textit{3},\textit{9}).

The GI also allows for the study of properties of specific regions; the enlarged clusters in \cref{fig:usecase}, which correspond to labels \textit{1}, \textit{5}, and \textit{7}, serve as examples.
The space between points in cluster \textit{1} is mostly stretched, contrary to the others; this can be explained by t-SNE's tendency to project all clusters at a similar density~\cite{wattenberg2016how}, and indicates that the embeddings of this cluster are actually more compact than what is being projected.
The opposite happens with cluster \textit{5}, which is shown to be more compressed than others, indicating that the embeddings of its elements were originally more dispersed.
Cluster \textit{7} presents both types of distortion, yet the GI is still useful for analyzing the cluster's shape and identifying which gaps are reliable and which have been distorted by the projection; this effect is also visible in cluster \textit{9}.

These observations not only prevent users from drawing na\"ive wrong conclusions about the data (e.g., assuming that all clusters are similar in density), but also provide valuable information to support analysis of global and local structure (e.g., which gaps between and within clusters are reliable).
Of course, the GI needs to be used in accordance with the analytical task at hand; if assessing the veracity of gaps is not a priority, other quality metrics might still be preferred.
As with other quality metrics, the GI is necessary to quantify some types of distortion, but should not be used as a single criterion to quantify the total quality~\cite{Machado_Behrisch_Telea_2025}.
Moreover, in order to avoid misusing the GI, users should also be aware of its limitations, which are discussed in the next section.

\section{Discussions and Limitations} \label{sec:discussion}

% Using GI on higher or lower dimension
\textbf{Using the GI beyond 2D projections.}
In this paper, we focus on the applicability of the GI to 2D layouts, which is the main use case for visual analysis of DR projections.
However, it should be noted that it can also be computed for projections of higher dimension, since the Delaunay triangulation generalizes to higher dimensions~\cite{fortune2017voronoi} (although with an increased computational cost), and the volume of an $n$-dimensional simplex can be computed through the Cayley–Menger determinant~\cite{kock2021}.
For one-dimensional projections, the notion of triangulation becomes degenerate; one could still consider segmenting the 1D layout into $1$-simplices (i.e., lines between adjacent points). Note that it would be different from stress, since the one-dimensional GI would only consider distances between adjacent points (i.e., gaps in the layout) rather than all possible pairs.
All these cases, although theoretically possible, require further study of the impact on the metric's interpretability, the types of distortion it could capture, and its visualization.

\begin{figure}[thb]
    \begin{subfigure}{0.4\columnwidth}
        \centering
        \includegraphics[width=0.28\linewidth,alt={Mock projection with 6 points aranged in a regular hexagon.}]{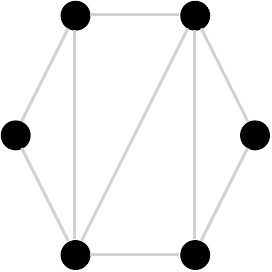} \\
        \vspace{1.5em}
        \includegraphics[width=\linewidth,alt={Mock projection with 6 points aranged in a hexagon stretched horizontally.}]{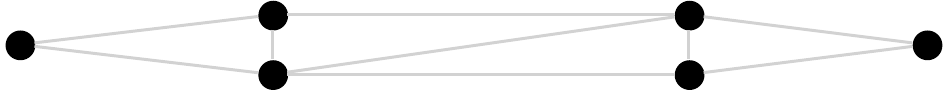} \\
        \vspace{0.5em}
        \caption{}
        \label{fig:limitations-fake-layouts}
    \end{subfigure}
    \hfill
    \begin{subfigure}{0.5\columnwidth}
        \includegraphics[width=0.49\linewidth,alt={GI visualization on a t-SNE layout. Some adjacent triangles have very different values.}]{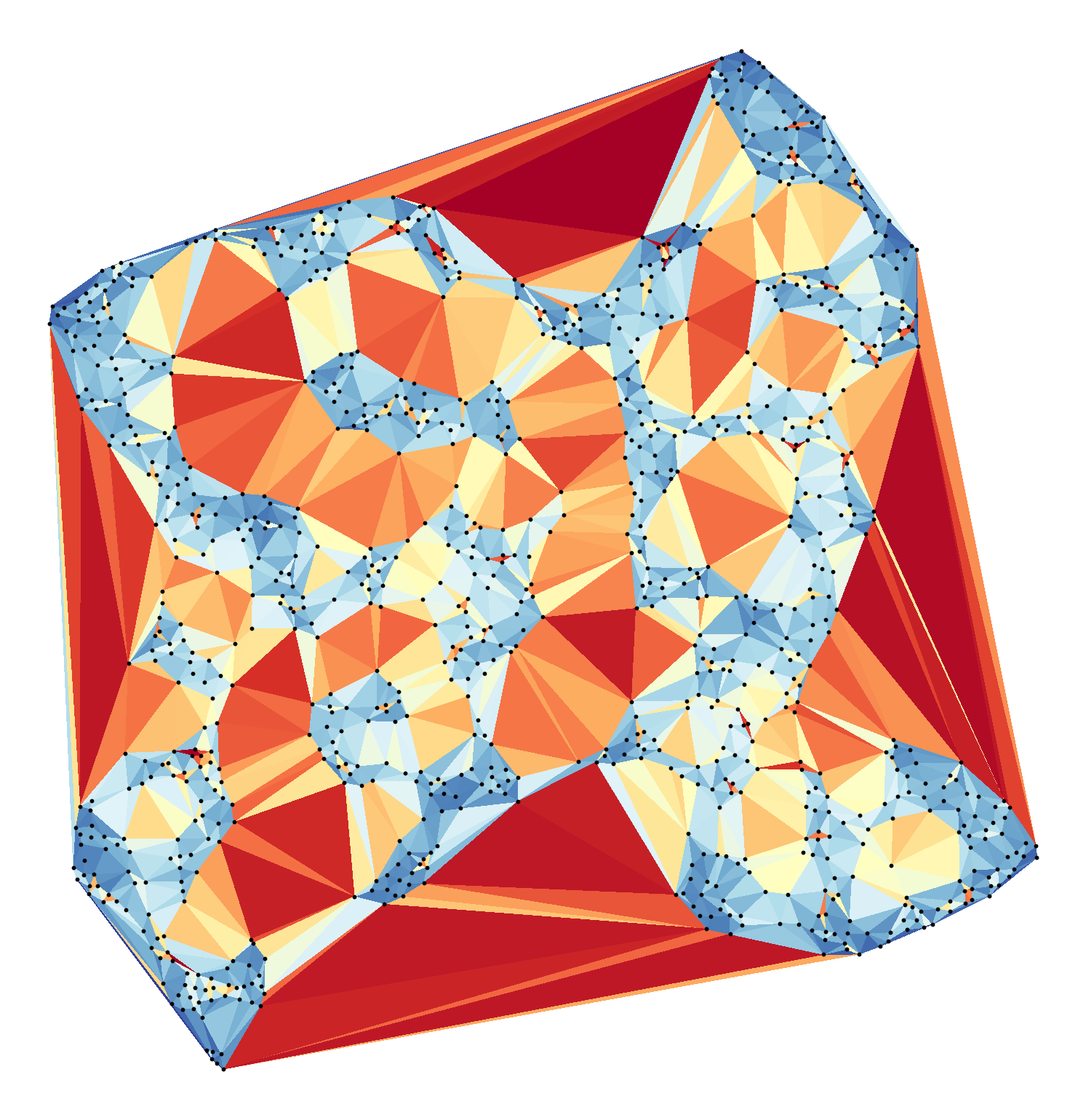}
        \includegraphics[width=0.49\linewidth,alt={GI visualization on a t-SNE layout, using the perimeter instead of area. Patterns are generally smoother, but the main deformations are consistent with the left figure.}]{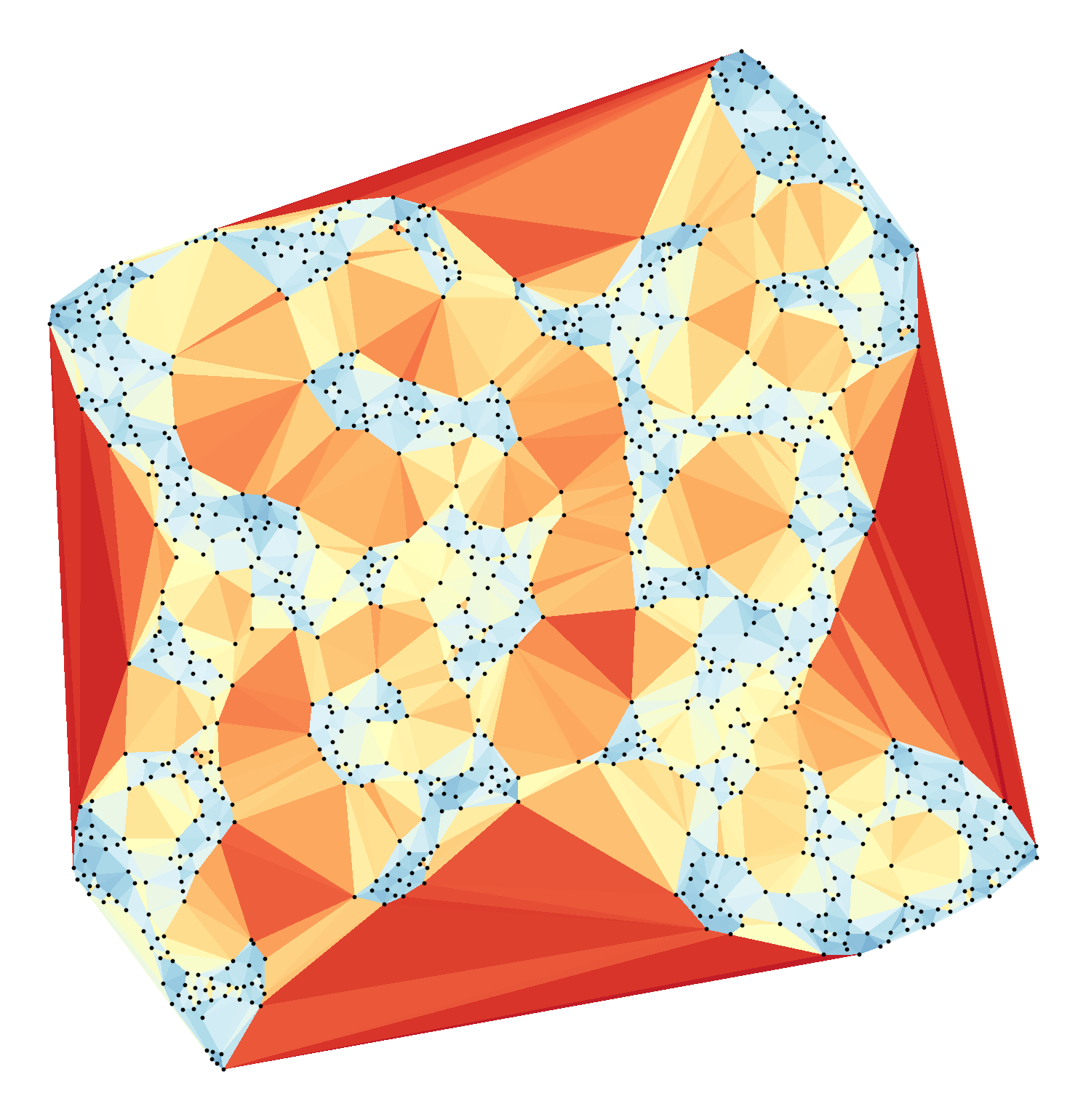}
        \caption{}
        \label{fig:limitations-perimeter}
    \end{subfigure}
    
    \caption{(a) Example of two 2D layouts whose triangulations (drawn for reference) generate triangles with the same relative area, but different perimeters. (b) t-SNE projections of the \textit{cube} dataset, visualized using the GI with deformations based on area (left) or perimeter (right).}
    \label{fig:limitations}
\end{figure}

% Deformations that cannot be captured with area
\vspace{0.25cm}
\noindent\textbf{Limitations of the triangle deformation approach.}
The triangle deformation function, defined in \cref{eq:deformation}, is based on the comparison of relative triangle areas. This makes the values easily interpretable as local compression and stretch (e.g., a distortion value of 0.9 indicates that the original relative area is only 10\% of the projected relative area).
However, while this has proven sufficient in the examples shown previously, there are cases where the position of the points can be distorted without changing the relative areas between them, such as uniform stretching of the whole layout, exemplified in the mock example in \cref{fig:limitations-fake-layouts}, which the current formulation of the GI cannot capture.

One can consider alternative deformation functions, which could potentially capture more complex types of distortion. In \cref{fig:limitations-perimeter}, we compare visualizations of the GI using relative areas and perimeters (additional examples are provided in the supplementary material). Although the results are consistent between the two approaches, the perimeter is less prone to having thin triangles with GI values different from those of their neighbors, resulting in smoother overall visualizations.
However, this comes at a cost of interpretability (the sum of perimeters depends on the specific triangulation and is meaningless in the context of the visual analysis) and stability (a change in triangulation will cause the sum of perimeters to be different, while the total area would remain constant).
Other more complex deformation functions, possibly involving angle comparison, could be explored in future work.%; in the current formulation of the GI, we strive for simplicity and interpretability.

Another limitation is the requirement for triangular distances in the high-dimensional space. Indeed, if the distances do not satisfy the triangular inequality, the computation area (\cref{eq:heron}) will be a complex value and the computation of the GI will fail.
We consider this to be a reasonable limitation, since the projection of a set of non-triangular distances is an ill-defined problem (e.g., what should a projection of three points $a$, $b$, and $c$, look like if their pairwise distances are not triangular?); for the same reason, other quality metrics such as stress are meaningless in these situations.

\begin{figure}[thbp]
    \centering
    \begin{subfigure}[c]{0.4\columnwidth}
        \includegraphics[height=3.5cm,keepaspectratio,alt={PCA projection of the Cube dataset using raw areas. The central area is yellow (no distortion), while the edges are dark blue (heavy compression).}]{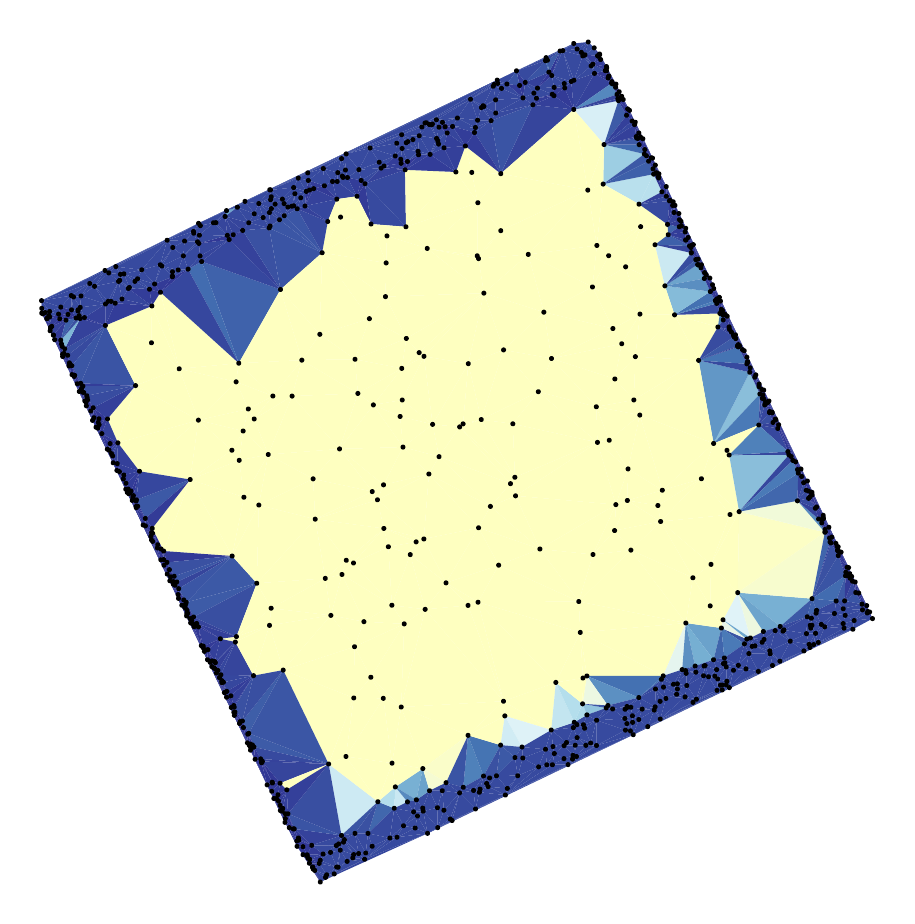}
        \caption{\textit{Cube} dataset -- PCA projection}
    \end{subfigure}
    \begin{subfigure}[c]{0.4\columnwidth}
        \includegraphics[height=3.5cm,keepaspectratio,alt={t-SNE projection of the plane dataset using raw areas. All triangles are colored dark red.}]{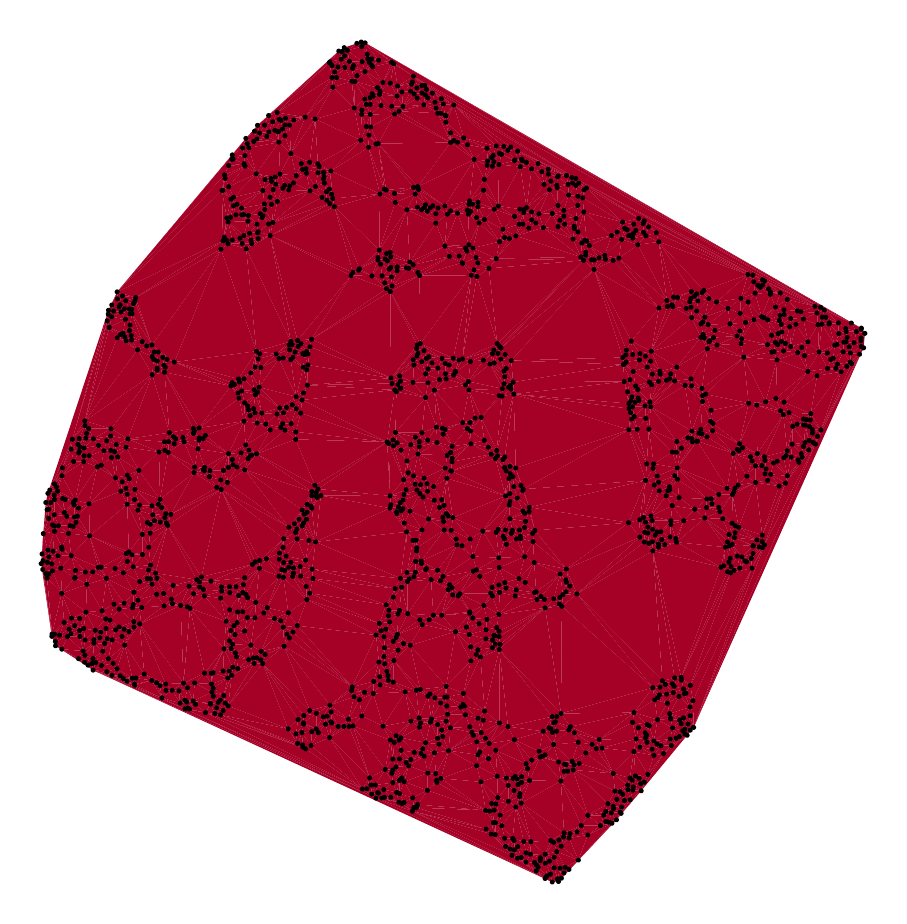}
        \caption{\textit{Plane} dataset -- t-SNE projection}
    \end{subfigure}
    \hspace{1em}
    \begin{subfigure}[c]{0.08\columnwidth}
        \includegraphics[height=3.5cm,keepaspectratio,alt={Colorbar blue-yellow-red.}]{figs/colorbar.pdf}
        \caption*{}
    \end{subfigure}
    \caption{Visualization of the GI using raw areas for two projections. Compare them to \cref{fig:results-interpreting-cube,fig:results-plane}b, respectively.}
    \label{fig:no-norm}
\end{figure}

% Relative areas
\vspace{0.25cm}
\noindent\textbf{Relative vs raw areas.}
The GI is intended to analyze visual distortion; thus, we focus on how important a gap in the layout is relative to all the others, also making the metric scale-invariant.
For this reason, in \cref{sec:met-triangulation}, we suggest computing the deformation based on the relative areas of triangles ($A'(t_i)$, $A'(\hat{t}_i)$), rather than the raw areas ($A(t_i)$, $A(\hat{t}_i)$).
To show the effect of this choice, in \cref{fig:no-norm}, we show two examples of scatterplots colored according to the triangle deformation with raw areas.% The examples correspond to the PCA projection of the cube with a missing face and the t-SNE projection of the \textit{plane} dataset.%; shown previously in Figures XX (left) and XX (right), respectively.

PCA, being an orthogonal projection of the data, will never stretch the raw area values; only compression is possible. Indeed, the middle area of the projection is perfectly preserved, while the other four sides of the cube are compressed.
Compare this to the default GI visualization, shown in \cref{fig:results-interpreting-cube}, where we consider the middle area to occupy disproportionately large space and to be stretched.
Thus, when using DR projection that preserves distances (such as PCA or mMDS), using raw areas is a reasonable alternative that can produce more intuitive results.

A limitation of using raw areas is that they are not scale-independent and can be meaningless when applied to projections that are not explicitly intended to preserve distances, such as t-SNE. Indeed, the t-SNE projection of the \textit{plane} dataset is shown as completely stretched, which does not provide any information as to where the visual distortion has occurred (compare it with the visualization in \cref{fig:results-plane}b).

% Alternatives to Delaunay
\vspace{0.25cm}
\noindent\textbf{Alternatives to the Delaunay triangulation.}
Prior to computing triangle deformations, one could consider alternative sets of triangles. In \cref{sec:met-triangulation}, we justified our choice of using the Delaunay triangulation, and its behavior has been validated in \cref{sec:results}.
But the GI can, in theory, be computed from any set of empty triangles in the projection, even if they overlap or do not cover the full 2D space.
A possible alternative is to compute the GI using all empty triangles in $\mathbf{X}$; however, this would be impractical to use beyond small datasets, since computing such a set has a cost linear to the number of empty triangles~\cite{dobkin_1988}, which is on the order of $\Theta(N^2)$~\cite{katchalski1988empty} (with a worst case of $\binom{N}{3}$, $N$ being the number of points).
%Moreover, such a set would be largely biased towards small (possibly very thin) triangles, since larger ones are less likely to be empty; one might consider filtering them based on elongation.
A small example of how such a triangulation would look, as well as some numerical results, is provided in the supplementary material.
Future work can explore more complex triangulation strategies, such as using empty triangles with certain properties of shape and size (e.g., discarding small or thin triangles) or computing triangles in high-dimensional space.

% Alternatives to visualization
\vspace{0.25cm}
\noindent\textbf{Open visualization challenges.}
Our suggested visualization approach assumes that the triangles occupy the full visual space (defined by the convex hull of the 2D layout) and do not overlap; this is guaranteed by the Delaunay triangulation, but using other sets of triangles, as suggested earlier, might require different visualization methods to handle transparency and other effects.
Moreover, the current triangle deformation function maps the triangle color to a single divergent normalized scale; if users choose to explore other deformation approaches, this should be adapted (e.g., using a 2D color space~\cite{Lespinats_Aupetit_2011} or showing different distortion types separately~\cite {Aupetit_2007}).

Although effective, the current simple approach can lead to the loss of important details, such as small areas in the scatterplot that affect the distortion.
One such case is heavily compressed areas, which are barely visible in the raw colored triangulation and could be erased through blurring. However, this problem is more general than just blurring, as compressed areas, despite having the same weight in the final aggregated result as stretched ones, are less visible in the scatterplot.

\vspace{0.25cm}
\noindent\textbf{Computing an embedding using the GI.}
Given a metric that computes the visual quality of an arbitrary embedding, a natural question is whether one can compute an embedding that optimizes such a metric, and what its properties would be.
Although the GI is fast to compute, its current form is not differentiable (due to the use of the Delaunay triangulation and the deformation and aggregation functions); one must then rely on derivative-free optimization methods, such as simulated annealing.
Moreover, edge cases should be considered, such as the limitation of the current deformation function to distinguish some types of distortion (\cref{fig:limitations-fake-layouts}), or degenerate cases like a 2D layout collapsed into a single line, where no triangles exist.
Given these complexities, we leave the design of a DR method based on the GI as future work.

\section{Conclusion} \label{sec:conclusion}

In this paper, we discussed the high perceptual impact of gaps in a 2D projection when visualized as a scatterplot, and the inability of standard quality metrics to meaningfully capture distortion in the layout's empty spaces.

To address this issue, we introduced the Gap Index (GI), a quality metric that measures the compression and stretch of empty regions in the projection. It does so by triangulating the 2D projection, relating each triangle to its high-dimensional counterpart to compute a triangle-distortion value, and finally aggregating all values into a single scalar.
We also suggested an approach to visualize local distortion directly in the scatterplot, allowing users to gain richer insights into the specific distortions present in the layout.
The running time of the GI is $O(N \log N)$, making it a scalable option to compute the quality of projections of large datasets.
Additionally, we presented quantitative and qualitative results to showcase its usage, compared it to other popular quality metrics, and analyzed its behavior.
In the discussion, we highlighted some limitations of the technique and suggested potential directions for future research.

%With the proposed modular structure of the GI, shown in Figure~\ref{fig:teaser} and discussed in depth in Section~\ref{sec:methodology}, we view the GI as an open method that can be modified, potentially targeting specific distortions to be captured. 
%However, the current results encourage us to recommend its use in daily practice for analyzing DR projections.

\bibliographystyle{abbrv-doi-hyperref}

\bibliography{references}

\end{document}